%% file: acl.tex
\pdfoutput=1
\documentclass[11pt]{article}

\usepackage{ACL2023}
\usepackage{times}
\usepackage{latexsym}

\usepackage[T1]{fontenc}
\usepackage[utf8]{inputenc}

\usepackage{microtype}

\usepackage{inconsolata}

\usepackage{graphicx}
\usepackage{amsmath}
\usepackage{xcolor}
\usepackage{appendix}
\usepackage{amssymb}
\usepackage{tabularx}
\usepackage{hyperref}
\usepackage{booktabs}
\usepackage{bbding}
\usepackage{multirow}
\definecolor{sample}{HTML}{6C8EBF} 
\definecolor{sample_projected}{HTML}{82B366} 
\definecolor{cylinder}{HTML}{005700}
\definecolor{green_back}{HTML}{CBEBC9}
\definecolor{blue_back}{HTML}{CAE8EC}
\definecolor{yellow_back}{HTML}{FFF2CC}

\usepackage{forest}
\usetikzlibrary{arrows.meta,shapes,positioning,shadows,trees}
\tikzset{
    basic/.style  = {draw, text width=2cm,  rectangle}, %drop shadow,= font=\sffamily,
    root/.style   = {basic, rounded corners=2pt, thin, align=center,
                     fill=green_back!60},
    onode/.style = {basic, thin, rounded corners=2pt, align=center, fill=blue_back!70,text width=2cm,},
    tnode/.style = {basic, thin, rounded corners=2pt, align=left, fill=yellow_back!80, text width=24em},
    xnode/.style = {basic, thin, rounded corners=2pt, align=center, fill=blue!20,text width=5cm,},
    wnode/.style = {basic, thin, align=left, fill=pink!10!blue!80!red!10, text width=6.5em},
    edge from parent/.style={draw=black, edge from parent fork right}
}
\title{Evaluating Readability and Faithfulness of Concept-based Explanations}

\author{
  Meng Li$^1$\footnotemark[1] , Haoran Jin$^2$\footnotemark[1] , Ruixuan Huang$^2$, Zhihao Xu$^1$, \\ \textbf{Defu Lian$^2$, Zijia Lin$^3$, Di Zhang$^3$, Xiting Wang$^{1}$\footnotemark[2]\textsuperscript{\dag}} \\
  $^1$ Renmin University of China \\
  $^2$ University of Science and Technology of China 
  $^3$ Kuaishou Technology \\
  }  
\begin{document}
\maketitle

\renewcommand{\thefootnote}{\fnsymbol{footnote}} %将脚注符号设置为fnsymbol类型，即特殊符号表示
\footnotetext[1]{These authors contributed equally to this work.} %对应脚注[1]
\footnotetext[2]{Corresponding author: \href{mailto:xitingwang@ruc.edu.cn}{\texttt{xitingwang@ruc.edu.cn}}} %对应脚注[2]

\renewcommand{\thefootnote}{\arabic{footnote}} %将脚注符号设置为fnsymbol类型，即特殊符号表示

\input{chapters/abstract}

\input{chapters/introduction}

\input{chapters/problem_formulation}

\input{chapters/methodology}

\input{chapters/experiments}

\input{chapters/conclusion}

\input{chapters/limitations}
\input{chapters/acknowledgements}
\input{chapters/ethical_statement}
% \section*{Acknowledgements}

% Entries for the entire Anthology, followed by custom entries
\bibliography{reference}
\bibliographystyle{acl_natbib}

\appendix

\input{chapters/appendix}

% 图片太大，导致compile超时，后期改成pdf格式优化一下，暂时先注释掉了，导致了一些warning

\end{document}

%% file: chapters/abstract.tex
\begin{abstract}

With the growing popularity of general-purpose Large Language Models (LLMs), comes a need for more global explanations of model behaviors. Concept-based explanations arise as a promising avenue for explaining high-level patterns learned by LLMs. Yet their evaluation poses unique challenges, especially due to their non-local nature and high dimensional representation in a model's hidden space. Current methods approach concepts from different perspectives, lacking a unified formalization. This makes evaluating the core measures of concepts, namely faithfulness or readability, challenging. To bridge the gap, we introduce a formal definition of concepts generalizing to diverse concept-based explanations' settings. Based on this, we quantify the faithfulness of a concept explanation via perturbation. We ensure adequate perturbation in the high-dimensional space for different concepts via an optimization problem. Readability is approximated via an automatic and deterministic measure, quantifying the coherence of patterns that maximally activate a concept while aligning with human understanding. Finally, based on measurement theory, we apply a meta-evaluation method for evaluating these measures, generalizable to other types of explanations or tasks as well. Extensive experimental analysis has been conducted to inform the selection of explanation evaluation measures. 
\footnote{Codes available at \url{https://github.com/hr-jin/Concept-Explanation-Evaluation}}

\end{abstract}

%% file: chapters/introduction.tex
\section{Introduction}

\begin{figure*}[htbp]
  \centering
  \vspace{-15pt}
  \includegraphics[width=\linewidth]{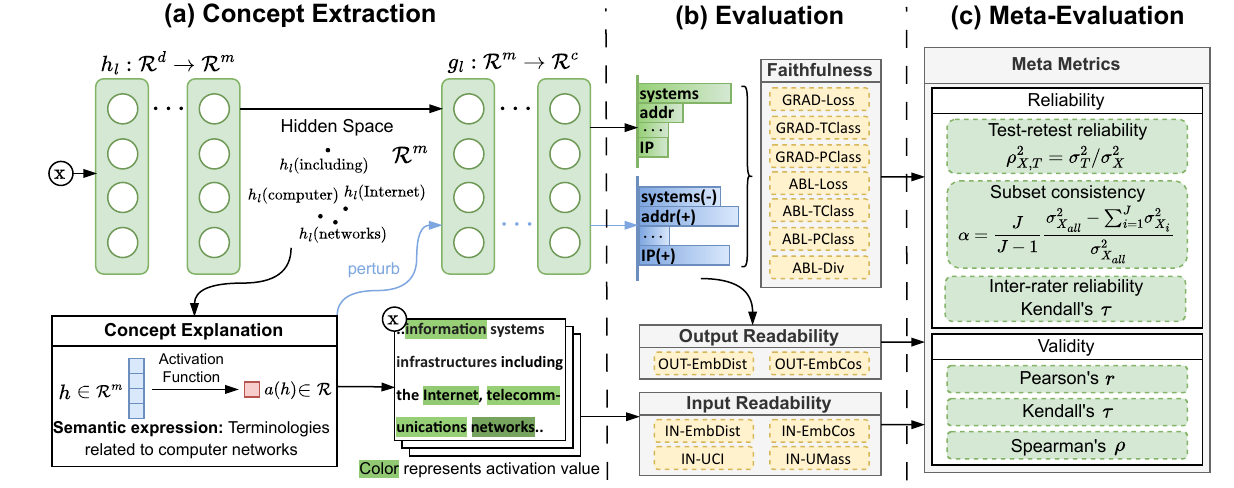}
  % \vspace{-20pt}
      % \caption{The overall framework of (a) concept extraction, (b) evaluation, and (c) meta-evaluation. (a) We formalize concepts as virtual neurons. (b) Evaluation is approached via readability and faithfulness. Readability is approximated by the semantic similarity of patterns that maximally activate the concept. Faithfulness is approximated by the difference in output when a concept is perturbed. (c) Meta-Evaluation is performed on the observed results of proposed measures via reliability and validity.}
      \caption{The overall framework. (a) Concept extraction: We formalize concepts as virtual neurons. (b) Evaluation is approached via readability and faithfulness. Readability is approximated by the semantic similarity of patterns that maximally activate the concept. Faithfulness is approximated by the difference in output when a concept is perturbed. (c) Meta-Evaluation is performed on the observed results of proposed measures via reliability and validity.\looseness=-1}
      \label{fig:framework}
      \vspace{-10pt}
\end{figure*}

Explainable Artificial Intelligence (XAI) holds significant value in pre-trained language models' mechanism understanding~\cite{li2022unified}, visualization~\cite{yang2024foundation}, performance enhancement~\cite{wu2023causal, ribeiro2016should,wang2022multi}, and security~\cite{burger2023your, zou2023representation}. Previous XAI algorithms have been applied to NLP tasks~\cite{wu2023causality}, vision tasks~\cite{wang2023densecl} and recommendation~\cite{jin2022towards,yang2022reinforcement}. These include natural language explanation~\cite{zhang2024distillation,lee2022self}, attention explanation~\cite{chen2019co,gao2019explainable}, and especially attribution methods~\cite{lundberg2017unified, sundararajan2017axiomatic,guan2019towards}. The attribution methods identify ``where'' the model looks rather than ``what'' it comprehends~\cite{colin2022cannot}, typically offering local explanations for a limited number of input samples, restricting their utility in practical settings~\cite{colin2022cannot,adebayo2018sanity}. \textbf{Concept}-based explanations~\cite{kim2018interpretability,cunningham2023sparse,fel2023craft} can mitigate the limitations of attribution methods by recognizing high-level~\cite{kim2018interpretability}  patterns (see Fig.~\ref{fig:framework}), which provide concise, human-understandable explanations of models' internal state.

% We evaluate the concepts' readability via highly activated fragments in input (left \textcolor{cylinder}{cylinder}), and highly enhanced output tokens (right \textcolor{cylinder}{cylinder}). Faithfulness is evaluated by difference between the output of \textcolor{sample}{original hidden representation} and \textcolor{sample_projected}{perturbed hidden representation}. 

%Despite these merits, the assessment of concept-based explanation methods presents significant challenges, unlike attribution methods. The absence of a definitive golden label for extracted concepts, coupled with the vast and often infinite array of possible concept sets~\cite{hoffman2018metrics}, i.e. countless directions in activation space shown as colored arrows in Figure~\ref{fig:framework}, complicates assessment. Without golden labels, user studies are also fraught with validation, standardization, consistency, and reproducibility issues~\cite{xiao2023evaluating}. In the meantime, current automatic metrics for evaluating concept-based explanations tend to be heuristic and method-specific~\cite{fel2023holistic}, impeding rigorous speculation and hindering the establishment of a standardized evaluation framework. 

% How good is any such concept in a network? A lack of a unified evaluation framework may hinder the development of such methods. Yet establishing this evaluation framework is never an easy task.
Despite these merits, the development of concept-based explanations may be hindered due to a lack of standardized and rigorous evaluation methodology. 
Unlike a single importance score assigned on each scalar input by attribution methods, diverse explanation methods approach high-dimensional concepts from different aspects. 
% endowed with high-dimensional concepts of methods with different principles vary.
This includes a single classification plane~\cite{kim2018interpretability}, an overcomplete set of basis~\cite{cunningham2023sparse}, or a module designed beforehand~\cite{koh2020concept}, lacking a unified landscape~(\emph{C1}).
% As concepts lie in a high-dimensional vector space in the middle of a DNN, prior faithfulness measures for traditional XAI methods~\cite{yeh2019fidelity,dasgupta2022framework}, which focus on scalar input elements, cannot be directly adopted. 
% The different perspectives taken by different design principles, as shown in Tab.~\ref{tab:baseline} further complicates assessment~(\emph{C1}). 
Moreover, its non-local nature across samples~\cite{kim2018interpretability}, combined with the high cost of human evaluation when the number of concepts is large, makes evaluating a concept's readability challenging~(\emph{C2}). For available evaluation measures~\cite{hoffman2018metrics}, it is difficult to test their reliability and validity~(\emph{C3}). 

In this paper, we address the challenges above and make the following contributions:

First, we \textbf{provide a unified definition of diverse concept-based explanation methods and quantify faithfulness under this formalization}~(\emph{C1}). By summarizing common patterns of concept-based explanation, we provide a formal definition of a concept, which can generalize to both supervised and unsupervised, post-hoc and interpretable-by-design methods, language and vision domains. Based on this, we quantify the faithfulness of a concept explanation via perturbation. We ensure adequate perturbation in the high-dimensional space for different concepts via an optimization problem.
% we view perturbations as an optimization problem and establish different degrees of faithfulness.

Second, we \textbf{approximate readability via coherence of patterns that maximally activates a concept}~(\emph{C2}). We utilize the formulation defined above to recognize patterns across samples that maximally activate a concept, from both the input and the output side. Then, we estimate how coherent they are as one concept via semantic similarity. Experimental results have shown this automatic measure correlates highly with human evaluation.\looseness=-1

Third, we \textbf{apply the classic measurement theory to perform a meta-evaluation on the faithfulness and readability measures}~(\emph{C3}). Measurement theory~\cite{allen2001introduction,xiao2023evaluating}  has been long utilized to verify whether a measurement is reliable and valid. Approaching via reliability and validity, this meta-evaluation method is useful for evaluating the measures for concepts and can be generalized to analyze the effectiveness of other measures, for example, measures for other types of explanations and other natural language tasks. Experimental results have filtered out 4 measures with low reliability, i.e. \textit{LLM-Score}, \textit{GRAD-Loss}, \textit{IN-UCI}, \textit{IN-UMass}, and verified the remaining faithfulness and readability measures’ validity.

%% file: chapters/problem_formulation.tex
\section{Concept Formalization}
\label{sec:problem_formulation}

% \subsection{Problem Formulation}
%We define the problem, illustrated in Figure~\ref{fig:framework}, as follows.
% [a summarization sentence]

% In this section, we present a formulation for the black-box models and the concepts.

%The problem we address involves understanding black-box large language models like GPT, which excel in text generation, a subset of classification tasks, making it easily extensible to other modalities.
% \textbf{Black-box model}.
% In our experiments, the black-box models to be explained are large language models (LLMs). Meanwhile, our method can be generalized to many other classification models, including image models.
In this paper, we primarily focus on explaining LLMs as black-box models. Meanwhile, our method can be generalized to many other deep classification models, including image models (see~Appx.\ref{app:applicability_image}).
As illustrated in Fig.~\ref{fig:framework}, we consider the black-box model to take an input $x$ from a dataset $D$ and output $y$, a $k$-class classification result. 
In text generation, $k$ is the vocabulary size.
% = |Vocab|$, where $|Vocab|$ is the vocabulary size. 
%Here, the input text $x$ originates from a dataset $D$, with $x^{t}$ denoting the $t$-th word.
%Here, the input text $x$ is from a dataset $D$.
%The model's output, $y$, represents a $k$-class classification result for each word. In text generation, $k = |Vocab|$, where $|Vocab|$ is the vocabulary size. 
For the $l$-th layer to be interpreted, given a sequence of input tokens $x^1, .... ,x^t$, their corresponding hidden representations are $h_l^1, ... ,h_l^t$.
% Considering the $l$-th layer of the model, its hidden space $\mathcal{H}_l$ is of dimension $m$, with the hidden representation of $x^{t}$ denoted as $h_l^t(x)$ and 
% The output logits for $k$ classes are $g_l(h_l^1),...,g_l(h_l^t)$. 
% The output logits for $k$ classes are $g_l(h_l^1,...,h_l^t)$. 
The output classification logits are $g(h)$. 

Within the context of Deep Neural Networks (DNNs), we summarize common patterns of concepts and establish a unified framework.
% We establish a unified framework for concepts within the context of DNNs. 
%Concepts, operating within the activation space of a DNN, serve as repositories of high-level knowledge, offering insights into the model's comprehension and decision-making mechanisms. 
Specifically, each concept is represented as a virtual neuron defined by an activation function that maps a hidden representation $h$ into a real value $a: \mathcal{R}^m \rightarrow \mathcal{R}$, where a positive output signifies activation. 
For each concept, a semantic expression may be given by humans or LLMs, depending on the concept explanation methods~\cite{kim2018interpretability,bills2023language}. Some methods take concepts and semantic expressions predefined by humans as inputs (e.g.,~\cite{kim2018interpretability}), while others require additional steps to produce a semantic expression based on highly activated tokens and samples of the extracted concepts~\cite{bills2023language}. Specifically, given high-activation samples of the concept and the highly activated tokens in these samples (e.g., ``Internet, computer, networks, …''), an LLM or a human labeler provides a semantic expression that summarizes their common patterns (e.g., ``Terminologies related to computer networks'')~\cite{bills2023language}.

\input{tables/baseline}

Our formalization can integrate diverse concept explanation methods, as shown in Tab.~\ref{tab:baseline}. This includes both supervised methods that require prior information about concepts (e.g., input samples that contain and do not contain the concepts)~\cite{kim2018interpretability} and unsupervised methods that do not rely on such prior information~\cite{ghorbani2019towards}. Our method also works for both post-hoc explanation methods that interpret a model after it is trained~\cite{kim2018interpretability} and interpretable-by-design approaches that integrate interpretability mechanisms directly into the model's architecture before training~\cite{koh2020concept}. Additionally, it applies to image backbone models as well.

%% file: tables/baseline.tex
\begin{table*}[htbp]
    \centering
    \begin{tabularx}{\textwidth}{cccc}
    \hline
        \toprule
         ~ & Method & Modal & Activation function $a(h)$ \\ 
        \midrule
        \multirow{4}{*}{Supervised} &  TCAV~\cite{kim2018interpretability} & text/image & $v^T h + b$  \\ 
        ~ & CBM*~\cite{koh2020concept} & image & $o_i^T h$ \\ 
         % \cline{2-5}   
         ~ & ProtoPNet*~\cite{chen2019looks} & image &  $\displaystyle \max_{\tilde{h} \in \mathrm{patches}(h)} \frac{\log ((||\tilde{h} - v||_2^2 + 1)}{||\tilde{h} - v||_2^2 + \epsilon}$\\ 
         \midrule
         \multirow{3}{*}{Unsupervised} & NetDissect~\cite{bau2017network} & image & $\displaystyle \frac{M(h) \cap L_c(x)}{M(h) \cup L_c(x)}$ \\ 
         & Neuron~\cite{bills2023language} & text/image & $o_i^T h$ \\ 
           & SAE~\cite{cunningham2023sparse} & text & $\mathrm{ReLU}(v^T h + b) $ \\ 
         \bottomrule
    \end{tabularx}
    \caption{Concept-based explanations' activation function. * denotes interpretable-by-design methods. Hyperparameters: 1) $v, o$ is a concept vector within the same space as $h$, and $o_i$ denotes a one-hot vector where $i$ indicates the position of the $1$ in the vector. 2) $M(h)$ selects the top-quantile activations and upsample them to the same dimension as $x$, and $L_c(x)$ is a pixel-level human-annotated label on $x$. 3) $b$ is a bias term.}
    \vspace{-10pt}
    \label{tab:baseline}
\end{table*}

% \begin{table}\centering \footnotesize
% \renewcommand{\arraystretch}{1.1}
%   \caption{Ablation study}
%   \vspace{-10pt}
%   \label{ablation-study-table}
%   \resizebox{\linewidth}{!}{
%     \begin{tabular}{clcccccc}
%     \hline
% & & \multicolumn{2}{c}{Beauty} & \multicolumn{2}{c}{Grocery} & \multicolumn{2}{c}{Health}  \\\cmidrule(lr){3-4}\cmidrule(lr){5-6}\cmidrule(lr){7-8}

% & & Overall & \OurMetric  & Overall & \OurMetric  & Overall & \OurMetric  \\
% \hline

% \multirow{4}{*}{NDCG} & Original & 0.2453 & 0.1900 & 0.2567 & 0.1443 & 0.2627 & 0.2378\\

% & Intra-GOOT & 0.2735 & 0.1601 & 0.2790 & 0.1244 & 0.2895 & 0.2072\\

% & Inter-GOOT & 0.2687 & 0.1590 & 0.2765 & 0.1281 & 0.2799 & 0.2188\\

% & \OurMethod & \textbf{0.2862} & \textbf{0.1495} & \textbf{0.2894} & \textbf{0.1119} & \textbf{0.3091} & \textbf{0.1915} \\ \hline

% \multirow{4}{*}{HR} & Original & 0.3752 & 0.2500 & 0.3925 & 0.2299 & 0.3755 & 0.2573\\

% & Intra-GOOT & 0.4199 & 0.2137 & 0.4244 & 0.1913 & 0.4248 & 0.2048\\

% & Inter-GOOT & 0.4095 & 0.2198 & 0.4201 & 0.1934 & 0.4165 & 0.2122\\

% & \OurMethod & \textbf{0.4331} & \textbf{0.1929} & \textbf{0.4432} & \textbf{0.1795} & \textbf{0.4477} & \textbf{0.1853} \\

% \hline 
    
%     \end{tabular}
%     }
% \vspace{-10pt}
% \end{table}

%% file: chapters/methodology.tex
%\section{Method}
%In this section, we present a systematic view of all metrics on concept evaluation and a rigorous analysis of the proposed metrics leveraging methods from measurement theory.

\section{Concept Evaluation Measures}
\label{sec:concept_evaluation}

% By surveying existing evaluation measures (Figure~\ref{fig:taxonomy} in Appendix~\ref{app:taxonomies}), we identify core concerns behind current measures: a concept's faithfulness to \textbf{machine} and readability to \textbf{humans}. 
We have conducted a literature survey on evaluation measures for concept-based explanations (Fig.~\ref{fig:taxonomy} in Appx.~\ref{app:taxonomies}), and decided to focus on two aspects that are of common interest: testing how well they reflect the underlying mechanisms of the machine (\textbf{faithfulness}) and assessing the extent to which explanations can be understood by humans (\textbf{readability}). \looseness=-1

\subsection{Faithfulness}

Widely studied in previous XAI methods, faithfulness is crucial for assessing how well a concept reflects a model's internal mechanism~\cite{chan2022comparative, lee2023neural,mccarthy1995faithfulness}.
However, its direct application to concept-based explanations presents challenges, 
particularly due to concepts' ambiguous representation in the hidden space of a model. The adequate degree of perturbation needed for diverse concepts extracted may vary, making it difficult to ensure a fair comparison.
% particularly due to the continuous nature of concepts. Unlike discrete input elements, faithfulness on continuous concepts is more complex, as we need to ensure the proper degree of perturbation.\looseness=-1

We quantify the faithfulness of a concept by the change in the output $g(h)$ after perturbing the hidden representation $h$ in the hidden space $\mathcal{H}$ where the concepts reside. We formulate faithfulness as $\gamma(a, \xi, \delta)$, where $\xi(h, a) $ applies a perturbation on $h$ given the activation function $a(h)$, and $\delta(y, y')$ measures the output difference. 
\begin{equation}
    \gamma(a, \xi, \delta)  =
     \frac{1}{|x|} \sum_{h^t \in f(x)}  \delta(y, y')
\end{equation}
with $y = g(h^t), y' = g(\xi(h^t, a))$ being the probability distribution of output vocabulary.

% \begin{equation}
%     \gamma(a, \xi, \delta)  =
%      \frac{1}{|h|} \sum_{h^t \in h}  \delta(y, y')
% \end{equation}
% where $y = g(h^t), y' = g(\xi(h^t, a))$ is a probability distribution of output vocabulary.

\textbf{Concept perturbation.} Based on the formalization of concepts in Sec.~\ref{sec:problem_formulation}, we view this problem as an optimization problem. As the concept formalization provided above encapsulates diverse kinds of concepts, this transformation allows the perturbation strategies to generalize beyond the linear form of concepts, like~\cite{chen2019looks}. 
% Here, we use $h'$ to denote the perturbed hidden representation.

Typical perturbation strategies include: 1) \textbf{$\xi_e$}: concept $\epsilon$-addition, wherein a near zero $\epsilon$ is introduced to maximally increase concept activation; 
2) \textbf{$\xi_a$}: concept ablation, which involves removing all information of the concept. 
The optimization problems can be formulated as:
% \begin{align}
%     \mathop{\arg\max}\limits_{h'} a(h'), & \quad \text{s.t.}\quad |h'-h| = \epsilon
%     \label{formula:optimization_epsilon}
%     \\
%     \mathop{\arg\min}\limits_{h'} ||h' - h||_2^2, & \quad \text{s.t.}\quad a(h) = 0
%     \label{formula:optimization_ablation}
% \end{align}
\begin{align}
    \xi_e(h, a) &= \mathop{\arg\max}\limits_{h'} a(h'), \ \ \ \text{s.t.} \ |h'-h| = \epsilon
    \label{formula:optimization_epsilon}
    \\
    \xi_a(h, a) &= \mathop{\arg\min}\limits_{h'}||h' - h||_2^2, \ \text{s.t.} \, a(h) = 0
    \label{formula:optimization_ablation}
\end{align}
% The closed form solution of Equation~\ref{formula:optimization_ablation} when $a(h) = v \cdot h$ is (please refer to Appx.~\ref{app:derivation} for detailed derivation):
% \begin{equation}
%     h' = h - \frac{v^Th}{v^Tv} v
% \end{equation}
% $\xi(h,a)$ takes the optimal $h'$. 
When the activation function is linear, e.g., $a(h) = v^T h$, the above problems have closed-form solutions (detailed derivation in Appx.~\ref{app:derivation}):
Correspondingly, the two perturbation strategies are: 
\begin{align}
% \text{(GRAD)} \quad   \xi_e(h, a) &= h + \epsilon v , \epsilon \rightarrow 0
\text{(GRAD)} \quad   \xi_e(h, a) &= \lim_{\epsilon \to 0} h + \epsilon v 
    \\
\text{(ABL)} \quad    \xi_a(h,a) &= h - \frac{v^Th}{v^Tv} v
\end{align}

\textbf{Output difference.} To quantify different aspects of faithfulness, we include i) difference in training loss ($\delta_l$), 
ii) deviation in logit statistics ($\delta_h$), 
iii) difference in the logit prediction of class $j$ ($\delta_c$), 
\begin{align}
 \text{(Loss)} \quad   \delta_l(y, y') & = \mathcal{L}(y, y^*) - \mathcal{L}(y', y^*)
    \\
 \text{(Div)} \quad   \delta_h(y, y') & = H(y, y')
    \\
 \text{(Class)} \quad   \delta_c^j(y, y') & = -(y^j -  y'^j) 
\end{align}
Here, $\mathcal{L}$ is a certain loss function~\cite{schwab2019cxplain,bricken2023monosemanticity}, $y, y'$ are the output classfication logits, $y^*$ is corresponding ground truth label, $y^j, y'^j$ are the logits of class $j$. To quantify the discrepancy between distributions, we utilize a statistic $H$, specifically KL-Divergence in our experimental setup.

For ease of reference, perturbations are expressed as prefixes, and difference measures are denoted as suffixes. Furthermore, we divide \textit{Class} into \textit{PClass} (prediction class) and \textit{TClass} (true class) with $j$ taking the predicted token class or ground truth token class. For instance, faithfulness computed via gradient to prediction class, as proposed in~\cite{kim2018interpretability}, is represented as \textit{GRAD-PClass}. Altogether, there are 2*4 kinds of available faithfulness measures. As the gradient option is too slow on vectors, we leave out \textit{GRAD-Div}.

\subsection{Readability}

% Readability assesses the extent to which humans can comprehend the extracted concept~\cite{lage2019evaluation}. Approaches such as sparsity and complexity, as demonstrated by~\cite{fel2023holistic, rosenfeld2021better}, aim to enhance readability by limiting the information presented to humans. Concurrently, studies like~\cite{dai2021knowledge, sajjad2022analyzing, do2019theory} delve into the relationship between extracted concepts and pre-defined, human-understandable labels. Another crucial facet of readability is separability~\cite{do2019theory, chen2020concept}, also known as purity~\cite{zarlenga2023towards}, diversity~\cite{vielhaben2022sparse}, or distinctness~\cite{ghandeharioun2021dissect}, emphasizing minimal intra-concept similarity.
% A human-understandable concept should only be activated on inputs that have a coherent meaning to humans, e.g., consistently activated when terminologies related to computer networks appear (Fig.~\ref{fig:framework}). 

% challenge: 1) non-local 2) human evaluation costly

Readability assesses the extent to which humans can comprehend the extracted concept~\cite{lage2019evaluation}. Most of the time, when patterns that maximally activate a concept are coherent (see example in Fig.~\ref{fig:framework}), can the concept be easily understandable to humans. 
We design coherence measures based on OpenAI’s pipeline~\cite{bills2023language} for human evaluation of concept quality. They presented human labelers with fragments where highly activated tokens were shown with color highlighting and asked the humans to try summarizing the commonalities of these highly activated tokens. We automate this process by assessing the commonality of highly activated tokens via co-occurrence or embedding similarity. 

As cross-sample patterns are extracted from a large corpus, diverse samples are needed to evaluate a concept's readability. Although previous efforts have made some progress in evaluating readability, they confront the challenge of ensuring data comprehensiveness while minimizing cost. Tab.~\ref{tab:readability_comparison} compares different measures for readability, including human evaluation~\cite{kim2018interpretability,ghorbani2019towards}, LLM-based measures~\cite{bills2023language,singh2023explaining}, and our proposed coherence-based measures. For the LLM-based evaluation, we considered~\cite{bills2023language,bricken2023monosemanticity}, which used less than 100 samples. For human evaluation, we considered the classical method by~\cite{ghorbani2019towards}, where each human rater scored no more than 20 samples per concept.

% Only concepts activated on a semantically coherent set of inputs, e.g. one that activates on only and all terminologies related to computer networks, are easily understandable to humans.
%We conduct a comprehensive literature study (Figure~\ref{fig:taxonomy} in the appendix), listing a comparison between different measures of readability in Table~\ref{tab:readability_comparison}.
% Tab.~\ref{tab:readability_comparison} compares different measures for readability, including human evaluation~\cite{kim2018interpretability,ghorbani2019towards}, LLM-based measures~\cite{bills2023language,singh2023explaining}, and our proposed coherency-based measures.
% We find current measures confront the challenge of ensuring comprehension while minimizing cost and subjectivity. Subjectivity denotes the degree to which the measure relies on individual interpretation or judgment. This may introduce potential variability in evaluations~\cite{honderich2005oxford}.\looseness=-1

\input{tables/readability}

\textbf{Human evaluation}.
Existing approaches predominantly rely on case studies and user studies~\cite{kim2018interpretability,ghorbani2019towards,chen2019looks}, asking humans to score a concept given a limited number of demonstrative samples. They are subject to issues of validation, standardization, and reproducibility~\cite{clark2021thats,howcroft2020twenty}. 

\textbf{LLM-based}.
As inexpensive human substitutes, LLMs have been utilized in evaluating concept-based explanations. A typical LLM-based score~\cite{bills2023language,singh2023explaining} is obtained by: 1) letting LLM summarize a natural language explanation $s$ for the concept (e.g., semantic expression in Fig.~\ref{fig:framework}) given formatted samples that maximally activates on the concept and activations $a$; 2) letting LLM guess the activation given only sample text and the generated explanation; 3) calculating an explanation score based on the variance between true activation and the simulated activation. However, the number of samples inputted to LLMs (4 in~\cite{bills2023language}) in step 1 is limited to maximum input length. This limits the comprehensiveness of the generated explanation, as shown in a case study in Appx.~\ref{app:case_study}. Even if the maximum input length is extended to 200k+ like Claude~3\footnote{\url{www.anthropic.com/news/claude-3-family}}, it may suffer from high computation cost and poor performance in long-dependency tasks~\cite{li2023loogle}.
% Despite their potential, LLM-based approaches have drawbacks, like relatively high cost (around \$1 for each concept on GPT4).\looseness=-1
%and low consistency, induced by the sampling operation. 
% We also find that current LLM-based evaluation measures often overlook readability from the output side and fail to yield stable results (see experiments in Sec.~\ref{sec:exp_reliability}). 
%and information about the activation function. %\looseness=-1
% the expression of concept explanations and fail to consider information
% Serving as a cost effective alternative of humans,  LLMs~\cite{bills2023language} present their drawbacks, including relatively high cost and low stability. Moreover, current evaluation metrics often overlook the expression of concept explanations and fail to consider information from the output side and activation function. 

\textbf{Coherence-based}.
To address these limitations, we propose novel measures inspired by topic coherence. Topic coherence measures are widely used in the field of topic modeling to estimate whether a topic identified from a large corpus can be easily understood by humans~\cite{newman2010automatic}.
% Similarly, \cite{bills2023language} have interpreted neurons automatically using LLMs by summarizing patterns that maximally activate a concept. 
Here, the basic idea is to approximate readability based on the semantic similarity between patterns that maximally activate a concept: we estimate how coherent they are as one topic (Fig.~\ref{fig:framework}). 
These measures mainly rely on the concept activation function, allowing for scalable, automatic, and deterministic evaluation.

% Beginning with a designated subset of texts, we feed them into a black-box LLM, extracting activations for the analyzed concept across each token. Subsequently, we extract tokens with high activation on the concept, utilizing both the token itself and its corresponding essential context to compute input-side readability. Moving beyond the input, we extend our evaluation to the output side by identifying tokens with the highest probabilities when setting the activation for the concept as $1$ and the rest as zero during the model's final prediction. To quantify readability, we employ semantic similarity measures, drawing on both traditional co-occurrence-based measures and deep embedding techniques.

Patterns that maximally activate a concept are obtained as follows. Initially, a subset of texts is selected and processed through a black-box LLM to obtain concept-specific activations for each token. High-activation tokens, indicative of a strong association with the analyzed concept, are then identified. For these tokens, important contextual words are extracted by ablating each word in the context and identifying those that impose the most impact on the high-activation token.
Similar information can be obtained from the output side. We extract tokens with the top-k highest likelihood when setting the hidden representation highly active on the concept and not on others. 

For our evaluation, we employ semantic similarity measures including \textit{UCI}~\cite{newman2009external}, \textit{UMass}~\cite{mimno2011optimizing}, and two deep measures \textit{Embedding Distance} (\textit{EmbDist}), \textit{Embedding Cosine Similarity} (\textit{EmbCos}). Each measure computes similarity $\mu(x^i,x^j)$ between two tokens $x^i,x^j$ as follows:
% For $coh \in \{UCI, UMass, ED, ES\}$:
% % 
% \begin{equation}
%     R(s, \mu_{coh}) = \frac{2}{|s|(|s|-1)}\sum_{0 \leq i<j < |s|} \mu_{coh}(s^i, s^j) \label{R_coh}
% \end{equation}
% % 
% Here, $s^{j}$ represents the $j$-th word in $s$, and $\mu_{PMI}$, $\mu_{UM}, \mu_{EmbDist}, \mu_{EmbCos}$ compute word correlation as follows:
%
% \begin{equation}
% 	\begin{split}
% 	\cos 2x &= \cos^2 x - \sin^2 x\\
% 	        &= 2\cos^2 x - 1
% 	\end{split}
% \end{equation}
\begin{align}
    % \begin{split}
    \mu_{\text{UCI}}(x^i,x^j) = &\log \frac{P(x^i,x^j) + \epsilon}{P(x^i) P(x^j)} 
    % \\
    % \mu_{NPMI}(x^i,x^j) = & \frac{\log \frac{P(w_i, w_j) + \epsilon}{P(w_i) P(w_j)}}{-\log(P(w_i, w_j) + \epsilon)} \label{mu_npmi}
    \\
    \mu_{\text{UMass}}(x^i,x^j) = &\log \frac{P(x^i,x^j) + \epsilon}{P(x^j)} 
    \\
    \mu_{\text{EmbDist}}(x^i,x^j) = &-|| e(x^i) - e(x^j) ||_2  
    \\
    \mu_{\text{EmbCos}}(x^i,x^j) = & \frac{e(x^i)\cdot e(x^j)}{|e(x^i)||e(x^j)|} 
    \label{formulation:coherence}
% \end{split}
\end{align}
Probabilities are estimated based on word occurrence frequency in the corpus. To prevent zero values in logarithmic operations, a small value $\epsilon$ is introduced. $e(x^i)$ embeds a word to a continuous semantic space, for example, using embedding models like BERT. 
% Due to the possibility of highly enhanced tokens not appearing in the dataset, the application of \textit{UCI} and \textit{UMass} measures is not viable for output labels.

For ease of reference and consistency, we denote readability on the input/output side using the prefixes \textit{IN/OUT}. For instance, readability computed using \textit{UCI} similarity on the input side is represented as \textit{IN-UCI}. Note that coherence-based measures may not capture all the desiderata of a readable explanation. Yet, it is still of interest to utilize this measure to filter a large amount of concepts when human evaluation may not be applicable.

\section{Meta Evaluation}
%\section{Evaluating Evaluation Measures}
\label{sec:meta_evaluation}

% contribution: reliability and validity, generalizability

How can we discern the effectiveness among possible measures available for evaluating concept-based explanations? 
% The potential absence of a golden label for extracted concepts, compounded by the vast and often infinite array of possible concept sets~\cite{hoffman2018metrics}, complicates assessment. 
Borrowing metrics from measurement theory~\cite{allen2001introduction} and psychometrics~\cite{wang2023evaluating,xiao2023evaluating}, our meta-evaluation focus centers on \textbf{reliability} and \textbf{validity}, guided by the methodological framework outlined in~\cite{allen2001introduction}. Our meta-evaluation methods can generalized to measures of a broader scope, including other XAI methods and other natural language tasks like generation.

% In our evaluation of concept-based explanations, we draw upon fundamental principles from measurement theory~\cite{allen2001introduction} and psychometrics~\cite{wang2023evaluating}, specifically focusing on reliability and validity, guided by the methodology presented in~\cite{xiao2023evaluating}.

\subsection{Reliability}

Reliability is crucial for assessing the consistency of a measure under multiple measurements, accounting for random errors introduced during measurement. These errors can arise from non-deterministic algorithms, data subsets, and human subjectivity. We particularly focus on three aspects: 1) \textbf{test-retest reliability}, quantifying the expected amount of uncertainty in the observed measure 2) \textbf{subset consistency}, measured as fluctuation across data subsets within a test; 3) \textbf{inter-rater reliability}, quantifying the degree of agreement between two or more raters.

%For a metric score $X$, with results $X_1, X_2$  from different test times, and the true score denoted as  $T$, we observe results $Y_1, Y_2, ... Y_J$  across $J$ subsets. The overall score on the entire dataset is expressed as $X_{all} = \sum_{i=1}^{J} Y_i$. \textit{Reliability stability} is calculated as the test-retest correlation $\rho_{X_1, X_2} = \rho^2_{X_1}/\rho^2_{X_2}$. \textit{Reliability consistency} is estimated through the lower bound $a$ of the coefficient of observed score and true score~\cite{cronbach1951coefficient}, as defined by the formula:
%\begin{equation}
 %   \rho^2_{X, T} \ge a = \frac{J}{J-1} \frac{\sigma^2_{X_{all}} - \sum_{i=1}^J \sigma^2_{Y_i}}{\sigma^2_X}
%\end{equation}
%Here, $\sigma^2_{X_{all}}$ represents the variance of $X_{all}$ across different models, and  $\sigma^2_{Y_i}$ denotes the variance of $Y_i$ across different models.

\textbf{Test-retest reliability} is quantified as the test-retest correlation: on the concepts extracted, we compute the same measure twice for each concept. The Pearson correlation~\cite{galton1877typical} between the two sets of results is test-retest reliability, which is an estimate of the expectation of:
\begin{equation}
\rho^2_{X, T} = \frac{\sigma^2_{T}}{\sigma^2_{X}}  
\end{equation}
where $X$ is the observed score, and $T$ is the true score, $\sigma_*^2$ denotes the variance of a random variable $*$. Typically, the minimal standard for an acceptable measure is 0.9~\cite{nunnally1994psychometric}.

\textbf{Subset consistency} is estimated through Cronbach's Alpha~\cite{cronbach1951coefficient}, a classic coefficient for evaluating internal consistency in measurement theory:
\vspace{-5pt}
\begin{equation}
    \alpha = \frac{J}{J-1} \frac{\sigma^2_{X_{all}} - \sum_{j=1}^J \sigma^2_{X_j}}{\sigma^2_{X_{all}}}
\end{equation}
$X_1, X_2, ... X_J$  are results of measure $X$ across different data subsets. The overall score on the entire dataset is expressed as $X_{all} = \sum_{j=1}^{J} X_j$.
$\alpha$ is the lower bound of squared correlation $\rho^2_{X, T}$ of observed score $X$ and true score $T$~\cite{cronbach1951coefficient}. For a measure with low subset consistency, one may use a larger test dataset to ensure the result's consistency.

\textbf{Inter-rater reliability} measures the degree of agreement across raters, calculated as score correlation among them. In this paper, we apply Kendall's $\tau$~\cite{kendall1938new} to measure pairwise correlation among raters using a scale that is ordered:
\begin{equation}
    \tau = \frac{2}{n(n-1)}\sum\limits_{i<j} \mathrm{sgn}{(X_i^1-X_j^1)}\mathrm{sgn}{(X_i^2-X_j^2)}
    \label{equation:kendall_tau}
\end{equation}
$X_i^*$ denotes the score on the $i$-th concept given by rater $*$. Evaluations that rely on humans must exhibit good inter-rater reliability, or, they are not reliable tests.

\subsection{Validity}

Validity is crucial in assessing how well a test measures the intended construct~\cite{nunnally1994psychometric}. A construct refers to the underlying criterion to be measured. In our case, it is faithfulness or readability. We focus on \textbf{concurrent validity},  evaluating the extent to which a test score predicts outcomes on a validated measure~\cite{cronbach1955construct}, and \textbf{construct validity}, examining how well indicators represent an unmeasurable concept~\cite{cronbach1955construct}. Construct validity can be further divided into convergent validity and divergent validity. 
% which assesses the correlation between theoretically related constructs, and \textbf{divergent validity}, which verifies a low correlation between theoretically unrelated measurements. 
% For example, in assessing IQ, we expect it to correlate with SAT and TOEFL scores (convergent validity) and expert scores (predictive validity) but not with measures of EQ (discriminant validity).

\textbf{Concurrent validity} reflects the appropriateness of a measure as an alternative to an existing reference, quantified via the correlation between the two scores. For example, an automatic measure for readability is used to approximate human evaluation at a large scale. Only when the automatic measure for readability is highly correlated with human scores, can we treat it as an approximate of human evaluation. Here we use classical correlation metrics to estimate concurrent validity~\cite{kendall1938new,spearman1961proof,galton1877typical}. Note that random error in either the automatic measure or human evaluation may impair concurrent validity. Thus being reliable is a premise of being valid.

\textbf{Convergent validity} verifies whether measures of the same construct are indeed related. For example, whether the purposed faithfulness measures are related to each other. As the underlying construct is often inaccessible to directly assess the measures' concurrent validity, convergent validity provides a statistic tool to assess construct validity via its relation~\cite{kendall1938new} with other measures of the same construct.

\textbf{Divergent validity} tests whether measures of unrelated constructs are indeed unrelated. For example, for distinct aspects considered of concept-based explanation (e.g., readability and faithfulness), measures of different aspects should show a significantly lower correlation than measures of the same aspect. Here we apply Kendall's $\tau$~\cite{kendall1938new} as a measure of correlation. A bad divergent validity may indicate potential bias in designed measures, calling for a more rigorous inspection of potential bias.

To inspect the construct validity of the measures to the intended constructs, we employ the multitrait-multimethod (MTMM) table methodology introduced by~\cite{campbell1959convergent}. This table conventionally presents pairwise correlations of observed measure scores on the off-diagonals and the subset consistency of each score on the diagonals. 

% For the choice of correlation measures, conventional correlations include Pearson's $r$~\cite{galton1877typical}, Kendall's $\tau$~\cite{kendall1938new}, or Spearman's $\rho$~\cite{spearman1961proof}, measuring pairwise correlation among raters using a scale that is ordered. Pearson assumes the rating scale is continuous; Kendall and Spearman's statistics assume only that it is ordinal.

%\looseness=-1 
% For our analysis, we adopt Kendall's $\tau$ to calculate correlation, as our measures are employed for comparing the ranked quality between concepts. Kendall's $\tau$ is chosen due to its suitability for ordered scales, follwing~\cite{xiao2023evaluating}.  The calculation of Kendall's tau is as follows: 
% \begin{equation}
%     \tau = \frac{2}{n(n-1)} \sum_{i<j} \text{sgn}{(x_i-x_j)} \text{sgn}{(y_i-y_j)}
% \end{equation}
% where $x_i, y_i$ are observations on the same data point among n data points computed via different measures, and sgn is the sign function.

%% file: tables/readability.tex
\begin{table}[htbp]
    \centering
    \footnotesize
    \vspace{-5pt}
    \begin{tabular}{cccc}
    \hline
        Method & \#Sample & Cost & Reliability   \\ 
    \hline
        Human & $< 20$ & high & medium  \\ 
        LLM-based & $<100$ & medium & low  \\ 
        Ours & $> 2000$ & low & high  \\ 
    \hline
    \end{tabular}
    \vspace{-5pt}
    \caption{Comparison of readability measures. \#Sample denotes the maximum number of samples applicable for evaluating a concept.}
    \label{tab:readability_comparison}
    \vspace{-10pt}
\end{table}

%% file: chapters/experiments.tex
\section{Experiments}

\subsection{Datasets and Experimental Settings}

We leverage the Pile dataset, a comprehensive collection curated by~\cite{gao2020pile}, which stands as the largest publicly available dataset for pre-training language models like Pythia~\cite{biderman2023pythia}. This dataset includes a vast 825 GiB of diverse data and encompasses 22 smaller, high-quality datasets spanning multilingual text and code. Its rich diversity facilitates the extraction of a wide array of concepts, crucial for our evaluation framework.

For the backbone model, we choose Pythia due to its pre-training on the Pile dataset, ensuring consistent knowledge representation between the training and explanation phases. 
% The Pythia family offers various model sizes and intermediate checkpoint results, enabling detailed analysis for nuanced insights. 
Additionally, we include GPT-2~\cite{radford2019language} to ensure the consistency of our findings across backbones (Appx.~\ref{app:sensitivity_analysis}). 
% GPT-2 is a key member of the GPT family, representing large-scale language models. 
Further details on these models are provided in Tab.~\ref{tab:model_property}.
To eliminate the impact of random fluctuations, we test each measure across 10 batches, each comprising 256 sentences with 128 tokens, totaling 327,680 tokens.

\subsection{Comparison of Evaluation Measures}
\label{sec:main_results}

In this section, we evaluate our proposed concept-based explanation measures, employing the meta-evaluation method for thorough assessment. 
To ensure a fair comparison, we randomly sampled 100 concepts extracted by each unsupervised baseline applicable to the language domain on the same backbone model, including Neuron-based method~\cite{bills2023language} and Sparse Autoencoder~\cite{cunningham2023sparse}.
% To ensure comprehensive coverage, we evaluate 100 randomly sampled concepts obtained through sparse dictionary learning and 100 randomly selected neurons treated as concepts. 
We primarily introduce results from the middle layer of Pythia-70M, with other consistent results across different layers and models in Appx.~\ref{app:sensitivity_analysis}.
Due to the possibility of highly enhanced tokens not appearing in the dataset, we apply \textit{UCI} and \textit{UMass} measures only on the input side.
% These are strategically extracted from the 3rd layer of the Pythia-70M model, balancing insights from both input and output spaces. Shallower layers offer human-comprehension-like insights, while deeper layers contain task-specific information. Considering concept diversity, 
% To eliminate the impact of random fluctuations, we test each measure across 10 batches, each comprising 256 sentences with 128 tokens, totaling 327,680 tokens.

% Besides measures proposed before, we add LLM simulation and score (LLM-Score)~\cite{bills2023language} for a more comprehensive analysis. 
% Due to new limitations in calculating logprobs on the input side, we adopt this \href{https://github.com/openai/automated-interpretability/blob/main/neuron-explainer/neuron_explainer/explanations/simulator.py#L638}{adjusted algorithm} with \texttt{gpt-4-turbo-preview} as the simulator.

\subsubsection{Reliability}
\label{sec:exp_reliability}
% [summary] 
%In this section, we analyze the retest reliability and subset consistency of automatic measures, as well as the inter-rater reliability of human evaluation.
In this section, we analyze which measures are reliable to random noise introduced by retesting, different data subsets, and human subjectivity.
%, i.e., how consistent their evaluation results are with respect to different choices of retesting, data subset consistency of automatic measures, as well as the inter-rater reliability of human evaluation.

\begin{figure}[htbp]
  \centering
  \vspace{-10pt}
  \includegraphics[width=0.95\linewidth]{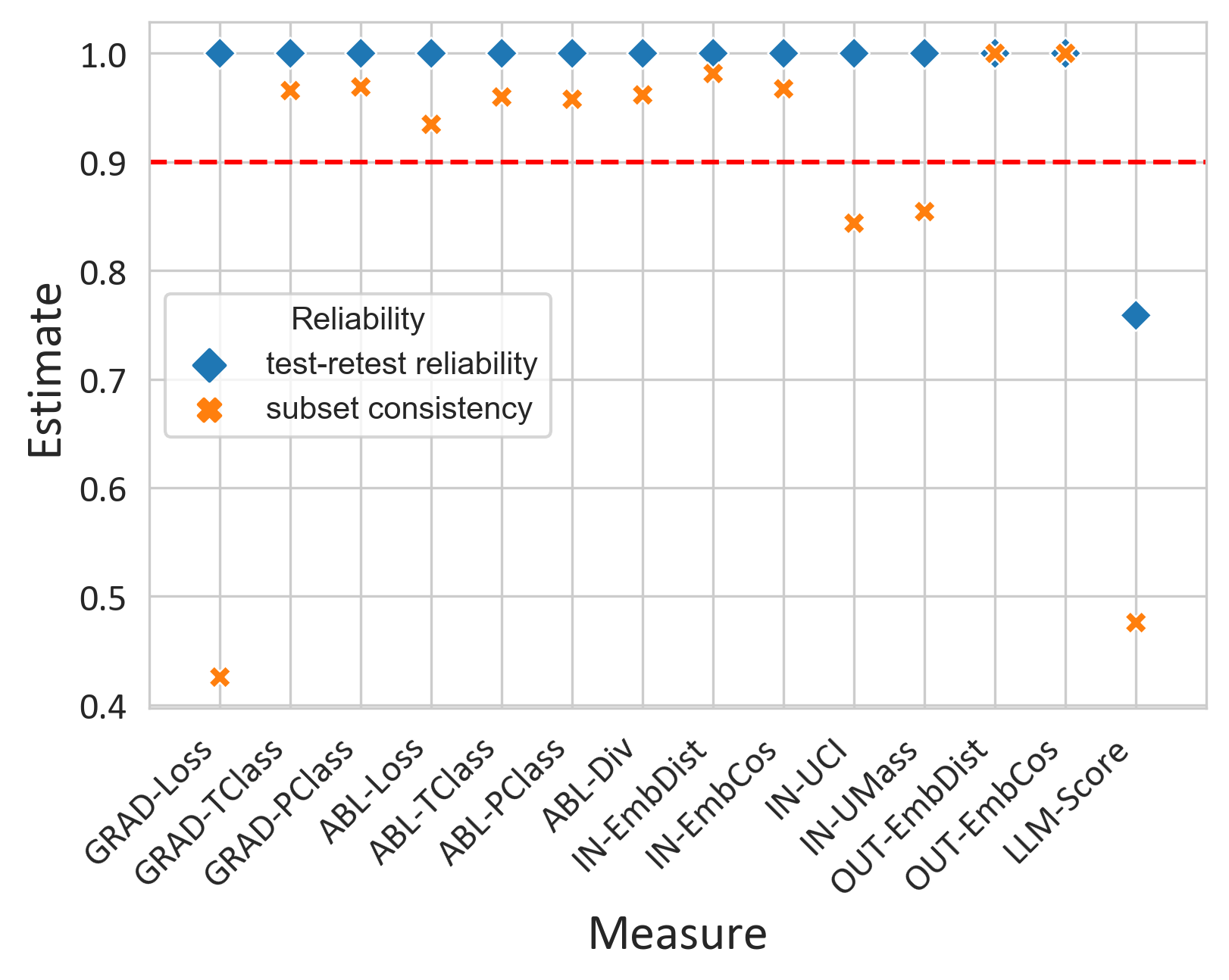}
  \vspace{-13pt}
      \caption{Estimated test-retest reliability and subset consistency of the proposed measures. The red dashed line indicates the minimal standard of 0.9~\cite{nunnally1994psychometric}.}
      \label{fig:reliability}
      \vspace{-10pt}
\end{figure}

%\noindent
\textbf{Test-retest reliability} results, depicted in Fig.~\ref{fig:reliability}, verifies the deterministic nature of the proposed measures, except for \textit{LLM-Score}~\cite{bills2023language}. \textit{LLM-Score} is less acceptable, which may be due to the inherent randomness introduced by sampling the most probable tokens.
% show that all of our proposed measures consistently received a perfect score of 1. This outcome underscores the deterministic nature of our measures, verifying their reliability across multiple tests with random seeds. In contrast, the LLM-Score exhibits lower reliability, less than the minimal standard of 0.9~\cite{nunnally1994psychometric}.
% This is due to the inherent randomness introduced during the random sampling among the most probable tokens to increase response diversity. Thus more careful design is needed when using LLMs as evaluator.

% \noindent\textbf{Subset consistency.} The diagnal of Figure~\ref{fig:mtmm_table_l3} illustrates the subset consistency results of the observed metrics, revealing that 10 out of 13 achieve a score exceeding 0.9, with only one lower than 0.8. There is a single outlier: GRAD-Loss. 
% Calculated via gradient computation, this metric exhibits high variance attributable to the inherent variability in gradient calculations.
% GRAD-Loss, measuring the contribution of a concept to loss, shows a low consistency across data subsets, highly subjecting to random error. This explains the fact that prior work mostly focus on GRAD-Class metrics, which are more consistent and meaningful. 
% Despite this outlier, the overall consistency of the metrics, with most scoring above 0.8, instills confidence in their reliability. With this assurance, we proceed to delve into further experimental analyses concerning validity of the measurements. 

\textbf{Subset consistency} provides further filtering of present measures with a threshold of 0.9~\cite{nunnally1994psychometric}, as shown in Fig.~\ref{fig:reliability}. For the faithfulness family, \textit{GRAD-Loss} shows an undesirable low consistency, probably due to the coupling of gradient and loss during training. For the readability family, \textit{IN-UCI} and \textit{IN-Umass} is less acceptable, attributing to the diverse nature of different concept's n-grams. Moreover, their capability to capture semantic similarity is also less desirable according to a case study shown in Appx.~\ref{app:case_study}

%In our examination of subset consistency, as depicted in the diagonal of Figure~\ref{fig:mtmm_table_l3}, the results show the performance of the observed measures. 
% The subset consistency results are presented in the diagonal of Figure~\ref{fig:mtmm_table_l3}.
% 10 out of the 14 measures achieved a score surpassing 0.9, demonstrating their high level of subset consistency. This collective reliability enhances our confidence in the measures, reassuringly indicating their resilience to random errors in the evaluation process.
% GRAD-Loss shows a low consistency of 0.43, probably due to their coupling during the training process.
% % The prevalence of prior work focusing on GRAD-Class measures instead of GRAD-Loss, elucidates the significance of this outlier. 
%  LLM-Score receives a lower subset consistency due to the randomness in explanation generation and simulation generation, coupled with relatively low retest reliability. Although it has been utilized in many prior work~\cite{cunningham2023sparse,bricken2023monosemanticity,palit2023towards}, we call for more careful consideration and correction before further use.
 %\looseness=-1

% \begin{figure}[htbp]
%   \centering
%   % \vspace{-10pt}
%   \includegraphics[width=\linewidth]{figures/stability.png}
%   % \vspace{-20pt}
%       \caption{Reliability consistency of the proposed metrics. A brighter color (higher value) indicates higher stability.}
%       \label{fig:stability}
%       \vspace{-15pt}
% \end{figure}

\textbf{Inter-rater reliability} is tested on human evaluation of readability.
% We conduct a user study on the readability of concepts and measure the inter-rater reliability for this human evaluation. 
%to further substantiate our proposed measures. 
%Following the recommendations outlined %in~\cite{jacovi2020towards}, we refrain from incorporating human judgment on the quality of interpretation or relying on human-provided gold labels for faithfulness evaluation. 
%Therefore, we focus our user study solely on assessing readability measures, which directly correlate with human comprehension. 
% Specifically, we curate a sample of 200 concepts, each comprising highly activated fragments on the input side and highly enhanced tokens on the output side. 
The concepts used for analysis above are scored by each human labeler with a high school level of English proficiency. They are blinded to the source method for the generated concepts and are tasked with scoring each concept on a scale of 1 to 5 based on two criteria: input readability and output readability. The recruitment of the experts and the setting of the user study are detailed in Appx.~\ref{app:user_instruction}.

\input{tables/expert_correlation}

%Here, we report the result of inter-rater correlation in Table~\ref{tab:user_corr} as a measure of reliability.
Tab.~\ref{tab:user_corr} shows the inter-rater reliability.
Overall, experts' correlations are high, with an average of 0.77 and 0.75 on the input and output sides.
%indicating high accordance among experts' results.

\subsubsection{Validity}
\label{sec:validity}

Here, we analyze whether the measures assess the intended construct, i.e., readability or faithfulness. We leave out \textit{LLM-Score}, \textit{GRAD-Loss}, \textit{IN-UCI}, \textit{IN-UMass} due to their low reliability as discovered in Sec.~\ref{sec:exp_reliability}. 
%effectivess assess readability or faithfulness
%the predictive validity of automatic measures, i.e., correlation with human evaluations, as well as construct validity i.e. correlation in-between measures.
%[summary]

\input{tables/user_study}

\begin{figure}[htbp]
  \centering
  % \vspace{-10pt}
  \includegraphics[width=\linewidth]{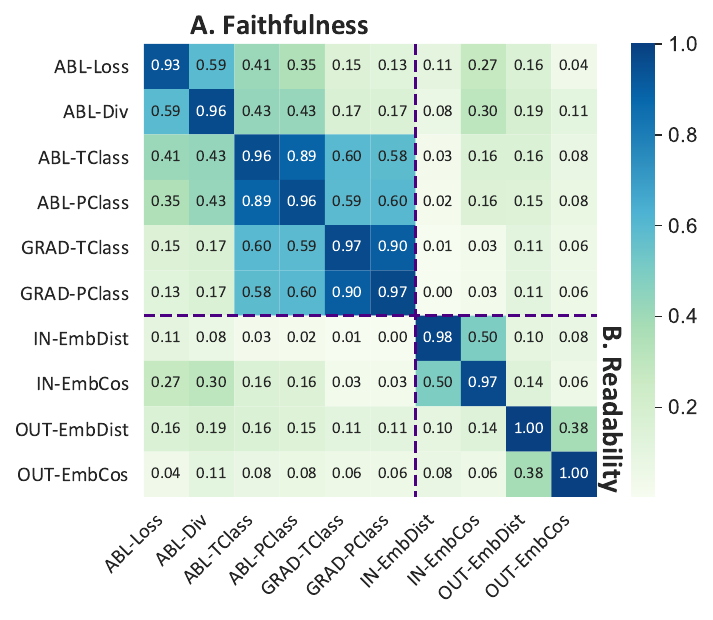}
  % \vspace{-20pt}
      \caption{
      %MTMM Table of the proposed measures. Construct validity is on the off-diagonals and subset consistency of each measure on the diagonals.
      The MTMM table of the evaluation measures: 1) subset consistency is shown on the diagonals; 2) construct validity is displayed on the off-diagonals.
      }
      \label{fig:mtmm_table_l3}
      \vspace{-10pt}
      
\end{figure}

\textbf{Concurrent validity.} 
In this experiment, we treat the user study results for readability as a criterion measure. Tab.~\ref{tab:user_study} shows how well existing automatic measures for readability correlate with user study results. %(predictive validity). 
% We have the following observations.
\textit{IN-EmbCos} is the top-performing measure to predict input readability (IR), and \textit{OUT-EmbCos} is the best in predicting output readability (OR). This demonstrates the effectiveness of our coherence-based measure EmbCos as an approximation of human evaluation. Compared with LLM-based measure that requires expensive API calls to GPT-4, EmbCos has a stronger correlation with human labels while requiring a much smaller computational cost. 
We recommend EmbCos as an inexpensive substitute for human evaluation, especially on large-scale evaluations. Yet human evaluation is still needed for more rigorous analysis.
In Fig.~\ref{fig:mtmm_table_l3}, the off-diagonals visually demonstrate the \textbf{construct validity} between our proposed measures. Our observations are as follows.

%revealing a notable distinction between faithfulness and readability metrics. Specifically,

\textbf{Divergent Validity} is inspected via correlation between unrelated measures. Measures of faithfulness (A) show a low correlation (0.0-0.3) with measures of readability (B), revealing their distinct nature, which is as expected. Input readability and output readability are also divergent (correlation less than 0.15), demonstrating concepts' unique patterns on both sides. While previous efforts on readability mostly focus on the input side, more careful inspection on the output side is needed.

\input{tables/readability_4cases.tex}

\textbf{Convergent Validity} is inspected via correlation between measures of the same construct. Faithfulness measures (A) displayed moderate correlations in general, averaging around 0.5. Agreement between measures with the same perturbation strategy or difference measurement is higher than others, indicating their potential relation. \textit{*-TClass} and \textit{*-PClass} showed a higher correlation, due to the consistency between prediction and true classes in well-trained language models. In the meantime, the agreement of readability measures (B) on either the input side or output side is moderate.

Our findings are consistent across different layers and backbones. Interested readers may refer to Appx.~\ref{app:sensitivity_analysis} for detailed results. 

% Notably, faithfulness and readability metrics capture distinct facets of these methods, evident in their low correlation. Within the faithfulness metrics, those belonging Abl family exhibit high correlations, indicating consistent evaluation within specific perturbation approaches. However, correlations decrease when comparing metrics across different perturbation strategies, reflecting the diverse perspectives from which concepts' effects are evaluated. However, the Grad strategy shows low correlations even within its family, coupled with low stability, raising concerns about its sensitivity to perturbation. 

% For readability metrics, distinctions between input and output sides are pronounced. Metrics within the same family on the input side share high relevance, while cross-family correlations are diminished. Traditional UCI/UMass struggle with synonym similarity due to word co-occurrence reliance, a challenge mitigated by EmbDist/EmbCos that encode words into continuous spaces, enabling more accurate modeling of semantic similarity. These findings underscore the nuanced considerations necessary when selecting and interpreting concept-based explanation metrics.

% \subsubsection{Sensitivity Analysis}

\subsection{Comparison of Explanation Methods}

\begin{figure}[htbp]
  \centering
  % \vspace{-10pt}
  \includegraphics[width=0.95\linewidth]{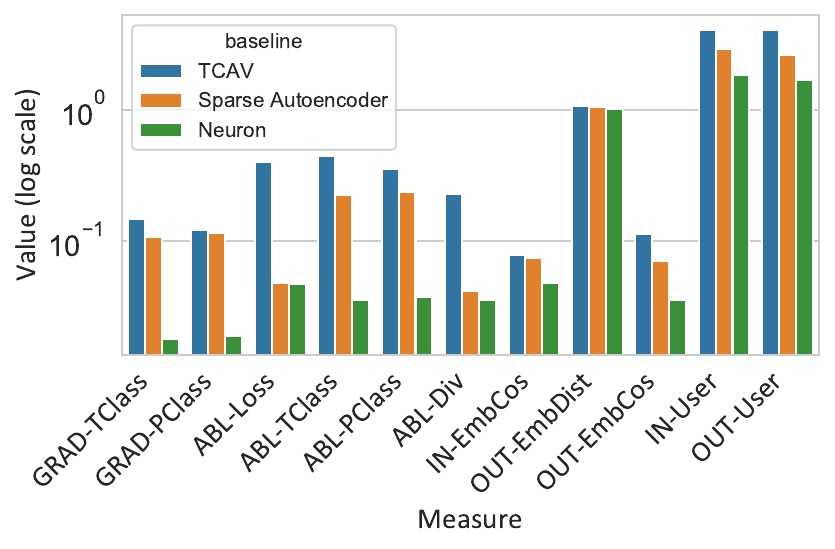}
  % \vspace{-8pt}
      \caption{Performance of different baselines on representative measures.}
      \label{fig:baseline_comparison}
      \vspace{-10pt}
\end{figure}

We conducted a comparative assessment of three different baseline methods on the language domain, including the concepts of neuron~\cite{bills2023language}, sparse autoencoder~\cite{cunningham2023sparse}, and TCAV~\cite{kim2018interpretability}. The results for both the neuron and sparse autoencoder were computed as the average values across 100 randomly sampled concepts from the concept set. We derive the supervised concept using TCAV following~\cite{kim2018interpretability,xu2024uncovering}. Initially, LLM's harmful QA~\cite{bhardwaj2023red} is treated as positive examples, and random texts are treated as negative examples. Their hidden representations are used to train a linear classifier, which aims to differentiate the representations of positive examples from negative ones. The trained classifier is treated as the concept’s activation function. The results of this analysis are shown in Fig.~\ref{fig:baseline_comparison}.

Sparse autoencoder surpasses the neuron-based methods across all evaluated measures, which is as expected. Nevertheless, as an unsupervised method, it falls short of TCAV on these same measures. This implies that the average quality of the concepts it extracted is not as high as the concepts derived from supervised counterparts. 
Additionally, the discrepancy between human ratings for different baseline methods is smaller than that between other readability measures. Upon detailed analysis of the results, it appears that human raters tend to give less discriminative scores ranging from 2 to 4, rarely awarding a 1 or 5, whereas automated measures show a greater range in scoring.

We also present a case study in Tab.~\ref{tab:readability_4cases} to visually illustrate the readability of concepts extracted by the three baselines. Firstly, TCAV's extracted concept shows high readability, with both input and output key tokens strongly tied to the ``harmful QA'' training theme. Secondly, The performance of the sparse autoencoder is notably inconsistent, whose concept set varies widely in readability measures. However, on average, upon observing many concepts, we found that the readability of concepts extracted by sparse autoencoder surpasses that of neurons. This suggests that the sparse dictionary paradigm generally enhances the quality of the entire concept set, mitigating the issue of superposition~\cite{elhage2022toy}.

Besides, we found that LLM has learned a seemingly redundant yet interesting pattern for the first concept shown for sparse autoencoder (e.g., north,·west, ·east, ·South, ·North, South, ·northern). Though these tokens are quite similar for humans, we do not know whether they are considered the same for LLMs. The embedding similarity between these tokens reflects LLMs’ ability to model them just like how humans perceive them as similar.

%% file: tables/expert_correlation.tex
\begin{table}[!ht]\footnotesize
    % \centering\footnotesize
    \vspace{-5pt}
    \setlength{\tabcolsep}{10pt}
    \begin{tabular}{cccc}
    % \toprule
    %      & Expert1 & Expert2 & Expert3  \\ 
    % \midrule
    %     Expert1 & - & 0.81 & 0.76 \\ 
    %     Expert2 & 0.77 & - & 0.74 \\ 
    %     Expert3 & 0.75 & 0.72 & - \\ 
    % \bottomrule
    \toprule
         & Input & Output & Average  \\ 
    \midrule
        Expert1\ \&\ Expert2 & 0.81 & 0.77 & 0.79 \\ 
        Expert1\ \&\ Expert3 & 0.76 & 0.75 & 0.76 \\ 
        Expert2\ \&\ Expert3 & 0.74 & 0.72 & 0.73 \\ 
    \bottomrule
    \end{tabular}
    \vspace{-5pt}
    \caption{Experts' Kendall's $\tau$ correlation as inter-rater reliability.}
    \label{tab:user_corr}
    \vspace{-10pt}
\end{table}

%% file: tables/user_study.tex
% \begin{table}[!ht]
%     \centering
%     \begin{tabular}{ccc}
%     \hline
%         Metric & IR & OR  \\ 
%     \hline
%         LLM Score & 0.164 & 0.416  \\ 
%     \hline
%         IN-EmbDist & 0.272 & 0.004  \\ 
%         IN-EmbCos & \textbf{0.483} & 0.039  \\ 
%         IN-UCI & -0.004 & -0.035  \\ 
%         IN-UMass & -0.072 & -0.084  \\ 
%         OUT-EmbDist & 0.147 & \textbf{0.446}  \\ 
%         OUT-EmbCos & 0.164 & 0.416  \\ 
%     \hline
%     \end{tabular}
%     \caption{Kendall $\tau$ correlation values for the different readability metrics with human scores: Input Readability (IR), and Output Readability (OR). The best performance is highlighted in bold.}
%     \label{tab:user_study}
%     \vspace{-10pt}
% \end{table}

\begin{table}[!ht]
    \setlength{\tabcolsep}{4pt}
    \centering
    \footnotesize
    % \tabcolsep=0.11cm
    % \vspace{-5pt}
    \begin{tabular}{ccccccc}
    \toprule
    
        ~ & \multicolumn{2}{c}{Kendall}  & \multicolumn{2}{c}{Pearson}  & \multicolumn{2}{c}{Spearman} \\ 
        \cmidrule(lr){2-3}\cmidrule(lr){4-5}\cmidrule(lr){6-7} 
        
        ~ & IR & OR & IR & OR & IR & OR \\ 
        \midrule

        LLM-Score & \underline{0.54} & 0.09 & \textbf{0.70} & 0.12 & \underline{0.67} & 0.12 \\ 
        \midrule
        IN-EmbDist & 0.19 & 0.12 & 0.27 & 0.16 & 0.26 & 0.16 \\ 
        IN-EmbCos & \textbf{0.56} & 0.18 & \underline{0.68} & 0.18 & \textbf{0.70} & 0.24 \\ 
        % IN-UCI & -0.02 & 0.05 & -0.09 & 0.09 & -0.04 & 0.07 \\ 
        % IN-UMass & -0.04 & 0.03 & -0.13 & 0.07 & -0.07 & 0.05 \\ 
        OUT-EmbDist & 0.15 & \underline{0.63} & 0.16 & \underline{0.73} & 0.21 & \underline{0.76} \\ 
        OUT-EmbCos & 0.17 & \textbf{0.67} & 0.16 & \textbf{0.75} & 0.23 & \textbf{0.80} \\

        \bottomrule
    \end{tabular}
    \caption{Concurrent validity of Input Readability (IR) and Output Readability (OR). The best results are marked in \textbf{bold}. The second-best results are \underline{underlined}.}
    \label{tab:user_study}
    \vspace{-10pt}
\end{table}

%% file: tables/readability_4cases.tex
\newcommand{\tabincell}[2]{\begin{tabular}{@{}#1@{}}#2\end{tabular}}

\begin{table*}[htbp]
    \centering
    
    \begin{tabularx}{\textwidth}{ccc}
    \hline
        \toprule
         Method & Input Relevant Tokens & Output Preferred Tokens \\ 
        \midrule
        TCAV &  {\tabincell{c}{
          $\cdot \text{information}$,  $\cdot \text{sensitive}$,  $\cdot \text{fraudulent}$,  \\ $\cdot \text{purposes}$,  $\cdot \text{violence}$,  $ \text{information}$, \\ $\cdot \text{candidate}$,  $\cdot \text{someone}$,  $\cdot \text{stealing}$,  $\cdot \text{hatred}$}} &  
          {\tabincell{c}{
          $\cdot \text{assassination}$,  $\cdot \text{illegal}$,  $\cdot \text{gren}$,$\cdot \text{rape}$, \\ $\cdot \text{unconstitutional}$,    $ \text{impeachment}$, $\cdot \text{/.}$, \\ $\cdot \text{prosecution}$,  $\cdot \text{unlawful}$,  $\cdot \text{conspiracy}$} }
         \\ 
         \hline
         
         \multirow{4}{*}{Sparse Autoencoder} &  {\tabincell{c}{
          $\cdot \text{north}$,  $\cdot \text{west}$,  $\cdot \text{east}$,  $\cdot \text{South}$, \\ $\cdot \text{North}$,  $ \text{South}$,  $\cdot \text{northern}$,  $\cdot \text{southern}$,  \\$\cdot \text{eastern}$,  $\cdot \text{dorsal}$}} &  
          {\tabincell{c}{
          $ \text{western}$,  $\text{ward}$,  $\text{bound}$,  $\text{side}$,  \\ $ \text{ampton}$, $ \text{wards}$,  $\cdot \text{facing}$, \\ $ \text{line}$,  $ \text{most}$,  $\cdot \text{coast}$} }
          \\
          \cline{2-3}  ~ & 
         {\tabincell{c}{
          $\cdot \text{task}$,  $\cdot \text{carbohydrates}$,  $\cdot \text{radiation}$,  \\ $\cdot \text{musician}$,  $\text{front}$,  $\cdot \text{version}$, \\ $ \text{own}$,  $\cdot \text{control}$,  $\cdot \text{Hope}$,  $\cdot \text{caution}$}} &  
          {\tabincell{c}{
          $\cdot \text{answer}$,  $\cdot \text{tumor}$,  $\cdot \text{disambiguation}$, \\ $\text{któ}$,   $\text{omitempty}$, $ \cdot \text{Version}$, \\ $\cdot \text{World}$, $\cdot \text{stream}$,  $\cdot \text{huh}$,  $\cdot \text{UK}$} }
          \\ 
         \hline
         Neuron &  {\tabincell{c}{
          $\cdot \text{gap}$,  $\cdot \text{als}$,  $\cdot \text{going}$,   $\cdot \text{3}$,  $\cdot \text{mit}$, \\ $\cdot \text{maybe}$,  $\cdot \text{True}$,  $\cdot \text{t}$,  $\cdot \text{c}$,  $\cdot \text{URN}$}} &  
          {\tabincell{c}{
          $ \text{lement}$,  $ \text{ters}$,  $ \text{right}$,$ \text{uki}$,  $\text{ter}$, \\   $ \text{ecycle}$, $ \text{aut}$, $\cdot \beta$,  $ \text{er}$,  $ \cdot \backslash \text{n}\backslash \text{n} $
          % \cdot \backslash \text{n}\backslash \text{n}} 
          }}
        \\
        \bottomrule
    \end{tabularx}
    \caption{Patterns that maximally activate some demonstrative concepts of the baselines. `$\cdot$' indicates space. For sparse autoencoder, we selected one concept from both the top 10\% and bottom 10\% based on the average rank results of \textit{IN-EmbCos} and \textit{OUT-EmbCos}. For the neuron method, we only showcased the top concepts.}
    \vspace{-10pt}
    \label{tab:readability_4cases}
\end{table*}

%% file: chapters/conclusion.tex
\section{Conclusion}
% This article has introduced standardized evaluation measures tailored specifically for concept-based explanations. These measures provide a universal framework that can be readily applied across a wide range of concept-based explanation methods, offering simplicity and accessibility without the need for additional human intervention. Furthermore, we have proposed a rigorous evaluation framework rooted in measurement theory principles, which enables the assessment of reliability and validity in these evaluation measures. By ensuring freedom from random errors and consistency across diverse settings, as well as freedom from systematic errors and accurate assessment of criteria, this framework enhances the robustness and effectiveness of concept-based explanation evaluation. Through extensive experimental analysis, we have validated and screened these evaluation measures, thereby facilitating the advancement and development of concept-based explanation methods.

This paper introduced two automatic evaluation measures, readability and faithfulness, for concept-based explanations. We first formalize a general definition of concepts and quantify faithfulness under this formalization. Then, we approximate readability via the coherence of patterns that maximally activate a concept. Another key contribution of this paper is that we describe a meta-evaluation method for evaluating the reliability and validity of these evaluation measures across diverse settings based on measurement theory. Through extensive experimental analysis, we inform the selection of explanation evaluation measures, hoping to advance the field of concept-based explanation.

% approaches evaluation measures tailored for concept-based explanations via readability and faithfulness, offering a unified framework applicable to various methods without human intervention. Based on measurement theory, we describe a method for evaluating the reliability, and validity of these evaluation measures across diverse settings, enhancing evaluation robustness. Through extensive experimental analysis, we inform the selection of explanation evaluation measures, advancing the field of concept-based explanation.

%% file: chapters/limitations.tex
\section*{Limitations}

Our framework may not encompass the entirety of the concept-based explanation landscape. Although the focus on readability and faithfulness aligns with prior research suggestions~\cite{jacovi2020towards,lage2019evaluation} and represents core components of evaluating concept-based explanations. We acknowledge that our study represents a modest step towards evaluating concept-based explanations. Future research on other aspects like robustness and stability is necessary.

Topic coherence is not designed to be the ultimate or perfect solution for measuring readability. Other aspects of readability, such as meaningfulness~\cite{ghorbani2019towards}, may also worth exploring. In the future, we are interested in investigating how these aspects could be quantified automatically, building a more comprehensive landscape of readability. 

Due to limited GPU resources and budget constraints, we used smaller versions of LLM, focusing primarily on the 3rd layer of Pythia-70M for our analysis. And our evaluation of the LLM-Score was restricted to 200 concepts, incurring a cost of around \$1 for a single concept. While this setup, on par with~\cite{cunningham2023sparse} and more general than~\cite{bricken2023monosemanticity}, allowed for fast analysis and comparison with existing literature, expanding our analysis to larger models could yield more insightful conclusions in the future.

% and potentila risk?
% required by ACL: Long papers may consist of up to eight (8) pages of content, plus up to one page for ethical considerations and/or limitations (see below), plus unlimited pages of references.

% \section*{Ethical Considerations}

%% file: chapters/acknowledgements.tex
\section*{Acknowledgements}
This work was supported by the National Natural Science Foundation of China (NSFC) (NO. 62476279), Major Innovation \& Planning Interdisciplinary Platform for the ``Double-First Class'' Initiative, Renmin University of China, Kuaishou, and the Fundamental Research Funds for the Central Universities, and the Research Funds of Renmin University of China No. 24XNKJ18. This work was partially done at Beijing Key Laboratory of Big Data Management and Analysis Methods and Engineering Research Center of Next-Generation Intelligent Search and Recommendation, Ministry of Education. This research was supported by Public Computing Cloud, Renmin University of China. 

%% file: chapters/ethical_statement.tex
\section*{Ethical Statements}

Our evaluation metrics for concept-based explanations offer a valuable contribution to enhancing human comprehension of LLM. However, it's crucial to acknowledge the potential presence of inherent hallucinations in the evaluation process that may have gone unnoticed.  

%% file: chapters/appendix.tex
% \section{Example Appendix}
% \label{sec:appendix}

% A list of symbols for reference:
% Basics
% \begin{itemize}
%     \item model $f(x): \mathcal{R}^d \rightarrow \mathcal{R}^m \rightarrow \mathcal{R}^k$
%     \item input $x$, the $t$-th word $x$
%     \item output $y$
%     \item dataset $D$
%     \item hidden space $\mathcal{H}_l\in R^m$
%     \item input -> hidden $h_l(x_t)$
%     \item hidden -> output $g_l(h_l(x_t))\in R^c$
%     \item concept $c_l^{(i)} = (v_l^{(i)}, s_l^{(i)})\in C_l$
%     \item concept vectors $V_l$
%     \item concept nlp representation $S_l$
% \end{itemize}
% Concept Extraction
% \begin{itemize}
%     \item concept extraction function $\phi(\mathcal{H}_l, D) = V_l$
%     \item identity matrix $E$
%     \item Cluster result $Cluster_i$
% \end{itemize}
% Concept Evaluation
% \begin{itemize}
%     \item extracting positive samples $\Omega(D, v)$
%     \item Readability $R(s, \mu)$
%     \item word similarity $\mu(x, x')$
%     \item faithfulness $\gamma(D, v, \xi, \delta)$
%     \item perturbation function $\xi(h, v)$
%     \item divergence measure $\delta(y, y')$
% \end{itemize}

\section{Taxonomies}
\label{app:taxonomies}

\input{figures/taxonomy}

In this section, we present a taxonomy of prior automatic measures for evaluating concept-based explanations based on existing literature on evaluating explainable AI~\cite{hoffman2018metrics,jacovi2020towards,colin2022cannot}. Fig.~\ref{fig:taxonomy} provides a summarized mind map, offering a visual representation of the various aspects by which concept-based explanation methods can be assessed. We endeavored to use the original terminologies as they appear in the cited works, emphasizing the purposes for which these measures were developed. 
Due to the evolving nature of the field, some measures might share similarities in their meanings or computational methods, which could lead to perceived overlap.
% However, it's important to note that the implementation of the same metric may vary across different papers, even in terms of concrete meaning, while different metrics across various papers may exhibit similarities. 

This makes the selection of suitable evaluation measures hard for practitioners in the field of concept-based explanation
Therefore, there is a pressing need for a more unified landscape in the evaluation of concept-based methods to facilitate substantial progress in the field. To address potential confusion, the evaluation measures we propose in this paper seek to clarify and distinguish between the different aspects of evaluation. We aim to provide a clear and structured approach that reflects the nuanced differences among these measures.

\section{Derivation of adequate ablation}
\label{app:derivation}

We consider concept ablation as an optimization problem with a closed-form solution, aiming to minimize perturbation while maintaining zero activation. This optimization problem can be formulated as:
\begin{equation}
    \mathop{\arg\min}\limits_{h'} ||h' - h||_2^2, \quad \text{s.t.}\quad \alpha(h) = 0
    \label{formula:optimization_app}
\end{equation}
We approach this optimization via the Lagrange multiplier.
For typical activation function calculated via inner product $\alpha(h) = v^T  h$, the Lagrange function is defined as:
\begin{align}
    \mathcal{L}(h, h',v) = ||h' - h||_2^2 + \lambda v^T  h'
\end{align}
On stationary points of $\mathcal{L}$:
% \begin{align}
%     \frac{\delta \mathcal{L}(h, h',v)}{\delta h'} &= 2(h'-h) + \lambda v
%     \\
%     \Leftrightarrow h' &= h - \frac{\lambda}{2} v
% \end{align}
\begin{align}
    &\frac{\delta \mathcal{L}(h, h',v)}{\delta h'} = 0
    \\
    \Leftrightarrow &2(h'-h) + \lambda v = 0
    \\
    \Leftrightarrow &h' = h - \frac{\lambda}{2} v
\end{align}
As $v^T h' = 0$, we have:
\begin{align}
    &v^T(h - \frac{\lambda}{2} v) = 0
    \\
    \Leftrightarrow & \lambda= \frac{2 v^T  h}{v^T v}
    \\
    \Leftrightarrow &h'= h - \frac{v^T h }{v^T v} v
\end{align}

For disentanglement-based methods, activation is calculated via $\alpha(h) = \mathrm{ReLU}(v^T h + b)$, where $\mathrm{ReLU}(x) = \max(x, 0)$. 
\begin{align}
    \text{When} \quad  & v^T h + b \le 0 
    \\
    \Leftrightarrow &\alpha(h) = 0
    \\
    \Leftrightarrow &h' = h
\end{align}
Otherwise, $\alpha(h) = v^T h + b$, the Lagrange function is defined as:
\begin{equation}
    \mathcal{L}(h, h',v) = ||h' - h||_2^2 + \lambda (v^T h' + b)
\end{equation}
On stationary points of $\mathcal{L}$:
% \begin{align}
%     \frac{\delta \mathcal{L}(h, h',v)}{\delta h'} &= 2(h'-h) + \lambda v
%     \\
%     \Leftrightarrow h' &= h - \frac{\lambda}{2} v
% \end{align}
\begin{align}
    &\frac{\delta \mathcal{L}(h, h',v)}{\delta h'} = 0
    \\
    \Leftrightarrow &2(h'-h) + \lambda v = 0
    \\
    \Leftrightarrow &h' = h - \frac{\lambda}{2} v
\end{align}
As $v^T h' + b = 0$, we have:
\begin{align}
    &v^T(h - \frac{\lambda}{2} v) + b = 0
    \\
    \Leftrightarrow & \lambda= \frac{2 (v^T h+b)}{v^T v}
    \\
    \Leftrightarrow &h'= h - \frac{v^T h +b}{v^T v} v
\end{align}

Similarly, we consider concept $\epsilon$-addition as an optimization problem with a closed-form solution, aiming to maximize concept activation with only perturbation of length $\epsilon$. This optimization problem can be formulated as:
\begin{equation}
    \mathop{\arg\max}\limits_{h'} a(h'), \quad \text{s.t.}\quad |h'-h| = \epsilon
    \label{formula:optimization_epsilon_app}
\end{equation}
The solution to this problem when activation function is $a(h) = v^T h$ is:
\begin{equation}
    h' = h + \frac{\epsilon}{|v|}v
\end{equation}

\section{Applicability to image domain}
\label{app:applicability_image}

In our paper, we mostly focus on LLMs as backbone models. Here we elaborate on how the proposed measures can be extended to the vision domain.

For readability, we can create `tokens’ by adopting a methodology similar to LIME~\cite{ribeiro2016should}. Specifically, we can segment each image into superpixels and regard each superpixel as a token in text. These superpixels' embeddings can then be obtained using pre-trained image models like VGG~\cite{simonyan2014very}, and coherence-based measures can be applied by assessing the similarity of these embeddings. While extending measures like \textit{UCI/UMass} to image tasks may present challenges, it remains feasible by first transcribing superpixels into text using vision-language models like CLIP~\cite{radford2021learning} and then calculating their co-occurrence. Yet considering the low reliability indicated in Sec.~\ref{sec:exp_reliability} as well as its original initiative for the language domain, it might be redundant to explore this variant.

Furthermore, faithfulness measures, operating on hidden and output spaces, are inherently independent of data modality and can be directly applied to image tasks. In general, our method can be used as long as a concept can be formulated with a virtual activation function (Sec.~\ref{sec:problem_formulation}), which takes a given hidden representation in the model as input and outputs the degree a concept is activated. As discussed in Sec.~\ref{sec:problem_formulation}, we believe this formulation is versatile and encompasses most concept explanation methods.

\section{Case Study}
\label{app:case_study}

\begin{figure*}[htbp]
	
	\begin{minipage}{0.32\linewidth}
		\vspace{3pt}

		\centerline{\includegraphics[width=\textwidth]{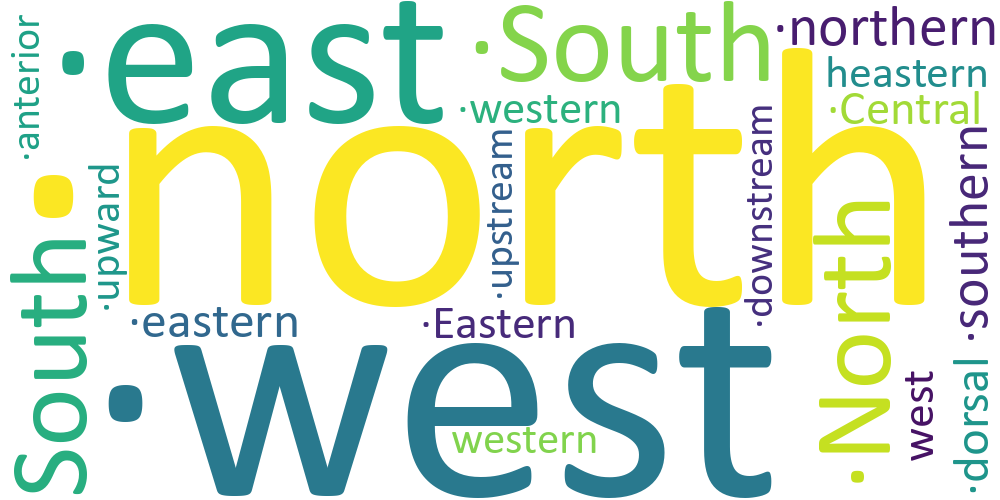}}

		\centerline{(a) Case 1}
        \label{fig:coherency_case1}
	\end{minipage}
	\begin{minipage}{0.32\linewidth}
		\vspace{3pt}
		\centerline{\includegraphics[width=\textwidth]{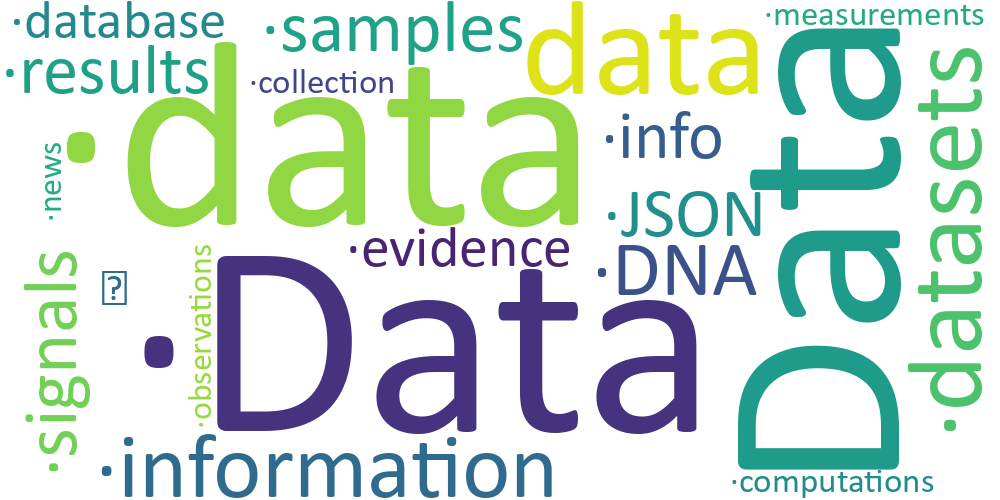}}
	 
		\centerline{(b) Case 2}
	\end{minipage}
	\begin{minipage}{0.32\linewidth}
		\vspace{3pt}
		\centerline{\includegraphics[width=\textwidth]{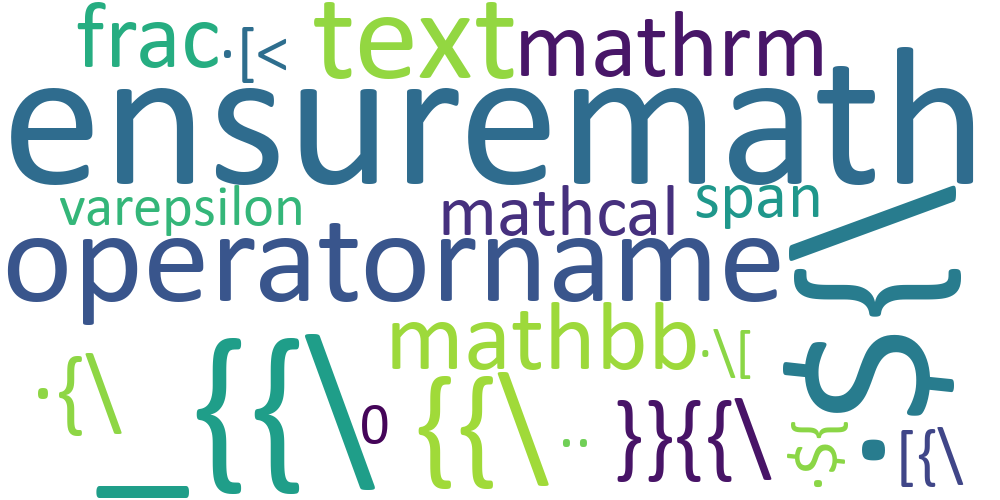}}
	 
		\centerline{(c) Case 3}
	\end{minipage}
 
	\caption{Topics extracted for calculating coherency-based measures. Spaces are replaced by `$\cdot$' for visualization. These topics align well with LLM-generated explanations in Fig.~\ref{fig:llm_case} while providing fine-granular information.  }
	\label{fig:coherency_case}
\end{figure*}

In this section, we present an illustrative case of the readability measures calculated via coherence-based measures and the LLM-based measure. We have the following observations.

\begin{figure}[htbp]
  \centering
  % \vspace{-15pt}
  \includegraphics[width=\linewidth]{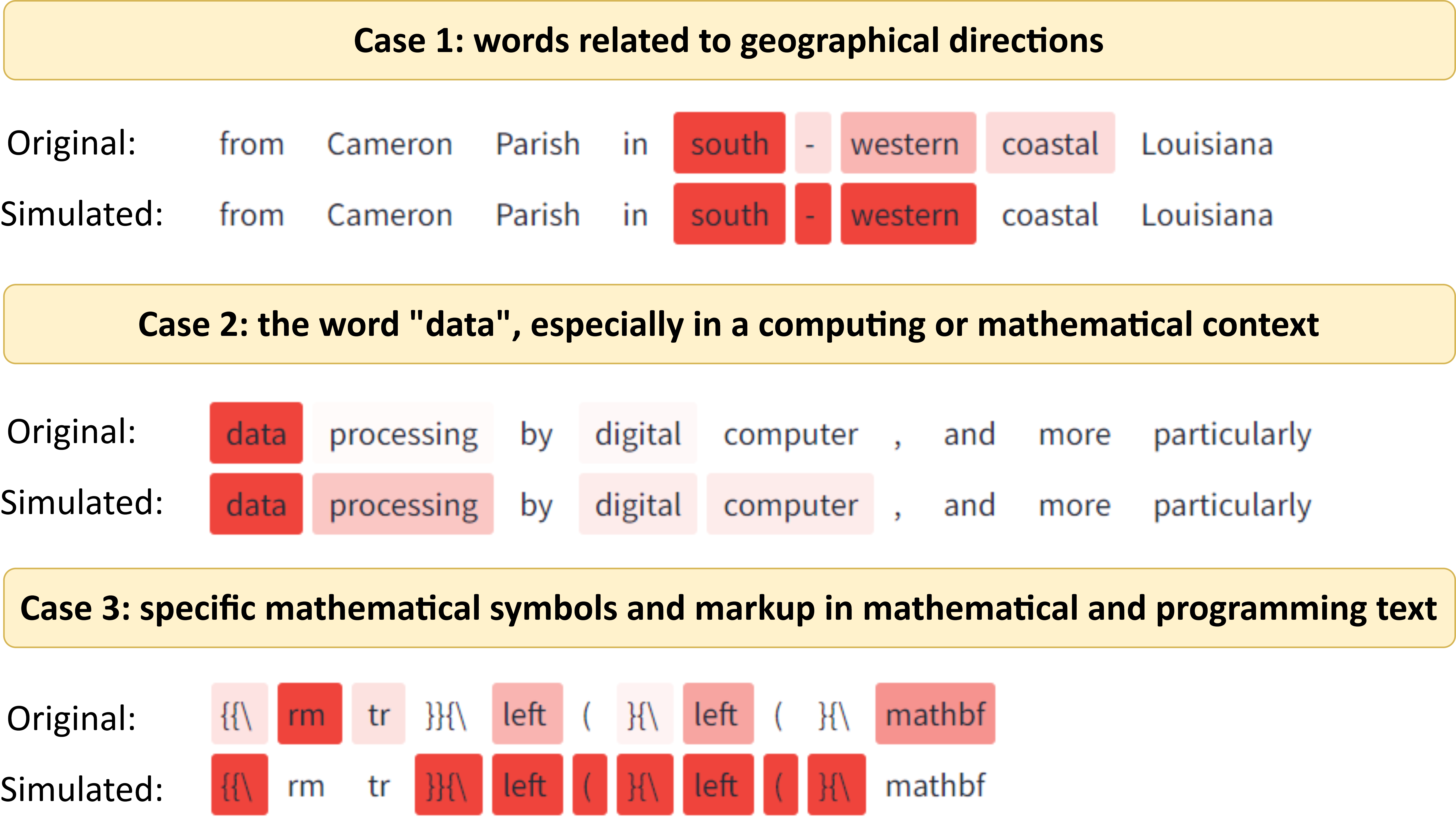}
  % \vspace{-20pt}
      \caption{A case study on LLM-based measure for readability measures. We present three cases with GPT4-generated explanation, original activation, and GPT4-simulated activation. GPT-4 performed well in the first two cases but worse in the third case.}
      \label{fig:llm_case}
      \vspace{-10pt}
\end{figure}

First, extracted topics via highly activated contexts align well with and even exceed explanations generated by LLM (Fig.~\ref{fig:llm_case}). As the number of samples inputted to LLM is restricted to a maximum context window and pricing limits (128,800 tokens and \$0.03/1K tokens for GPT-4), explanations generated by LLM are only limited to the information presented. However, our coherency-based measures can search from a broader range of samples, looking for top-activating contexts to provide a more comprehensive explanation, as shown in Fig.~\ref{fig:coherency_case}.

Second, deep embedding-based measures are better at capturing semantic similarities. The first case illustrated in Fig.~\ref{fig:coherency_case} (a) is ranked as the 1st among the 200 concepts evaluated by \textit{IN-EmbCos} and  3rd by LLM, as it consistently activates on words related to geographical directions as suggested by LLM. However, \textit{IN-UCI} only assigned a rank of 172. This is largely attributed to the fact that these terms may only occur once in a sample, showing one single direction, leading to low word co-occurrence counts.

Third, coherency-based measures can compensate for failure cases of LLM. For the 3rd case shown in Fig.~\ref{fig:coherency_case}, we can observe that it activates on expressions related to \LaTeX. However, as LLM can only observe limited examples, it fails to include other attributes than mathematical symbols and markup, thus failing to simulate activations that align with the original activation. We approach this challenge by extracting topics from a larger range of samples.

Overall, these findings are consistent with the ones disclosed in Sec.~\ref{sec:validity}, offering a more intuitive understanding of the measures' advantages and weaknesses.

\section{Sensitivity Analysis}
\label{app:sensitivity_analysis}

% \begin{figure}[htbp]
%   \centering
%   % \vspace{-15pt}
%   \includegraphics[width=\linewidth]{figures/mtmm_l1.pdf}
%   % \vspace{-20pt}
%       \caption{The MTMM table of the evaluation measures on the 1st layer of Pythia-70M: 1) subset consistency is shown on the diagonals; 2) construct validity is displayed on the off-diagonals.}
%       \label{fig:mtmm_l1}
%       \vspace{-10pt}
% \end{figure}

\begin{figure*}[htbp]
	
	\begin{minipage}{0.33\linewidth}
		\vspace{3pt}

		\centerline{\includegraphics[width=\textwidth]{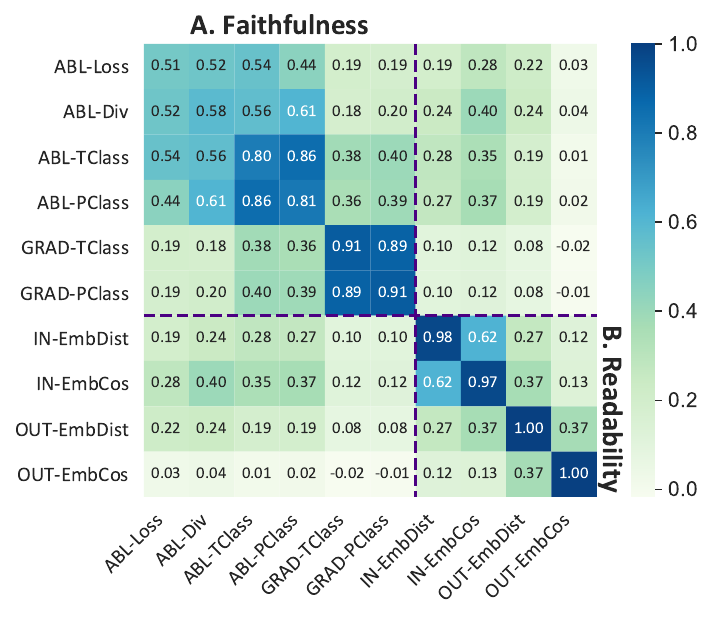}}

		\centerline{(a) 1st layer of Pythia-70M}
        % \label{fig:mtmm_l1}
	\end{minipage}
	\begin{minipage}{0.33\linewidth}
		\vspace{3pt}
		\centerline{\includegraphics[width=\textwidth]{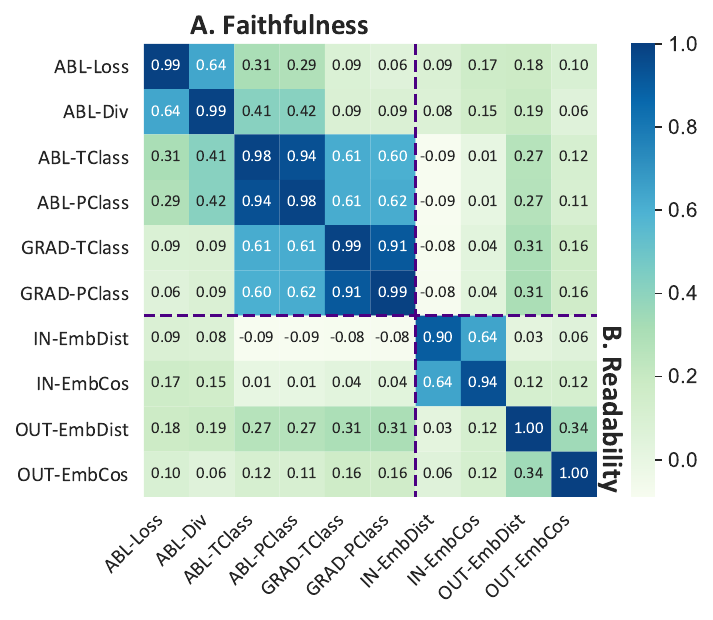}}
	 
		\centerline{(b) 5th layer of Pythia-70M}
        % \label{fig:mtmm_l5}
	\end{minipage}
	\begin{minipage}{0.33\linewidth}
		\vspace{3pt}
		\centerline{\includegraphics[width=\textwidth]{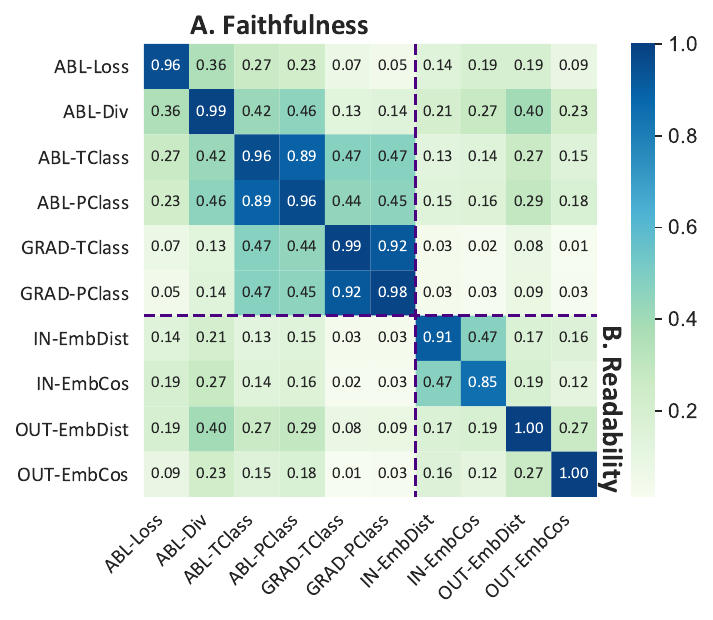}}
	 
		\centerline{(c) 6th layer of GPT-2 small}
        % \label{fig:mtmm_gpt2_l6}
	\end{minipage}
 
	\caption{The MTMM table of the evaluation measures: 1) subset consistency is shown on the diagonals; 2) construct validity is displayed on the off-diagonals.}
	\label{fig:sensitivity_analysis}
\end{figure*}

In our sensitivity analysis of validity results, we expand beyond the examination of the 3rd layer of Pythia-70M, as depicted in Fig.~\ref{fig:mtmm_table_l3}. We include results from the 1st (Fig.~\ref{fig:sensitivity_analysis} (a)) and 5th (Fig.~\ref{fig:sensitivity_analysis} (b)) layers of Pythia-70M, as well as results from the 6th layer of GPT2-small (Fig.~\ref{fig:sensitivity_analysis} (c)). Across these layers, reliability and validity results are consistent, with measures showing slightly better subset consistency in deeper layers. We speculate that as the layers deepen, the model discards irrelevant information and noise, leading to more stable and robust representations that are subject to less random error and exhibit higher consistency. Notably, the validity results on the 6th layer of GPT2-small align with our main findings (Fig.~\ref{fig:sensitivity_analysis} (c)), fluctuating within a reasonable range, typically less than 0.1. These results underscore larger language models' superior ability and reliability compared to their counterparts, such as the 3rd layer of Pythia-70M.

% \begin{figure}[htbp]
%   \centering
%   % \vspace{-15pt}
%   \includegraphics[width=\linewidth]{figures/mtmm_l5.pdf}
%   % \vspace{-20pt}
%       \caption{The MTMM table of the evaluation measures on the 6th layer of Pythia-70M: 1) subset consistency is shown on the diagonals; 2) construct validity is displayed on the off-diagonals.
% }
%       \label{fig:mtmm_l5}
%       \vspace{-10pt}
% \end{figure}

% \begin{figure}[htbp]
%   \centering
%   % \vspace{-15pt}
%   \includegraphics[width=\linewidth]{figures/mtmm_gpt2.pdf}
%   % \vspace{-20pt}
%       \caption{The MTMM table of the evaluation measures on the 6th layer of GPT-2 small: 1) subset consistency is shown on the diagonals; 2) construct validity is displayed on the off-diagonals.
% }
%       \label{fig:mtmm_gpt2_l6}
%       \vspace{-10pt}
% \end{figure}

\section{Implementation Details}
\label{app:implementation}

In our implementation, we employ the Pile dataset, truncating input to 1024 tokens for efficient analysis. Both the \href{https://huggingface.co/datasets/EleutherAI/pile-standard-pythia-preshuffled}{Pile dataset} and the backbones utilized are accessible for download from the Hugging Face Hub. To compute embedding-based readability measures, we leverage the backbone model's embedding matrix to extract token embeddings. All correlation metrics utilized in our analysis are calculated using the \texttt{scipy} package.

\input{tables/model_stat}

% We appreciate your attention to details regarding the experimental settings and implementation details. We will improve the experimental settings in Section 4.1 and implementation details in Appendix C. Specifically, we will add the details regarding neuron-based methods in a paragraph after line 1030 with the following content: ‘Following Bills et al.[13], Cunningham et al. [14], we adopt the extraction of neuron activation as the output of the MLP layer in each layer, where each dimension corresponds to a neuron. Specifically, for a feed-forward layer $FFN(h) = GeLU(h W_1) W_2$, the MLP output/neurons are $GeLU(hW_1)$. Furthermore, the disentanglement-based baseline can utilize these extracted neurons as inputs to discover mono-semantic concepts, leveraging sparse autoencoders.’ To note the original source of baselines, we will add these contents in line 1046: ‘It's important to note that our approach is in line with the original experimental setups outlined in Bills et al.[13], Cunningham et al. [14], and for a more detailed understanding of the implementation settings, interested readers are encouraged to refer to the original papers.’

Following~\cite{bills2023language,cunningham2023sparse}, we adopt the extraction of neuron activation as the output of the MLP layer in each layer, where each dimension corresponds to a neuron. Specifically, for a feed-forward layer $\mathrm{FFN}(h_{in}) = \mathrm{GeLU}(h_{in} W_1) W_2$, the MLP output/neurons are $\mathrm{GeLU}(h_{in} W_1)$. Furthermore, the disentanglement-based baseline can utilize these extracted neurons as inputs to discover mono-semantic concepts, leveraging sparse autoencoders. We obtain the concept activation function of TCAV following~\cite{kim2018interpretability}. We treat LLM's harmful QA~\cite{bhardwaj2023red} as positive examples, and random texts as negative examples. Then, a linear classifier is trained on their hidden representations to classify harmful examples. The trained classifier's output function is regarded as the concept’s activation function. 

We employ a sparse autoencoder proposed by~\cite{cunningham2023sparse} to obtain concepts for the disentanglement-based baseline. The process involves running the model to be interpreted over the text while caching and saving activations at a specific layer, as narrated above. These activations then constitute a dataset used for training the autoencoders. The training is executed with the Adam optimizer, employing a learning rate of 1e-3, and processing 11 billion activation vectors for one epoch. The dictionary size is set at 8 times the hidden space's dimension. A single training run with this data volume is completed in approximately 13 hours on a single RTX 3090 GPU. To balance between sparsity and accuracy, we set the coefficient on the L1 loss to 0.5 for the 3rd layer of Pythia-70M.

It's important to note that our approach is in line with the original experimental setups outlined in~\cite{bills2023language,cunningham2023sparse,kim2018interpretability}. For a more detailed understanding of the implementation settings, interested readers are encouraged to refer to the original papers.

In calculating faithfulness, GRAD-Div is neglected as gradient operation is only applicable to one variable at a time, applying gradient operation to the whole output class is computationally expensive. To aggregate the effect on each token, they are weighted by their activations. Samples that exhibit high activation levels regarding a specific concept are deemed more relevant to the concept empirically and therefore receive higher weights. This weighting scheme ensures that the most representative samples contribute more significantly to the evaluation, enhancing the fidelity of the faithfulness measure in capturing the alignment between the model's behavior and the intended concept. 

\begin{figure*}[htbp]
  \centering
  % \vspace{-15pt}
  \includegraphics[width=0.7\linewidth]{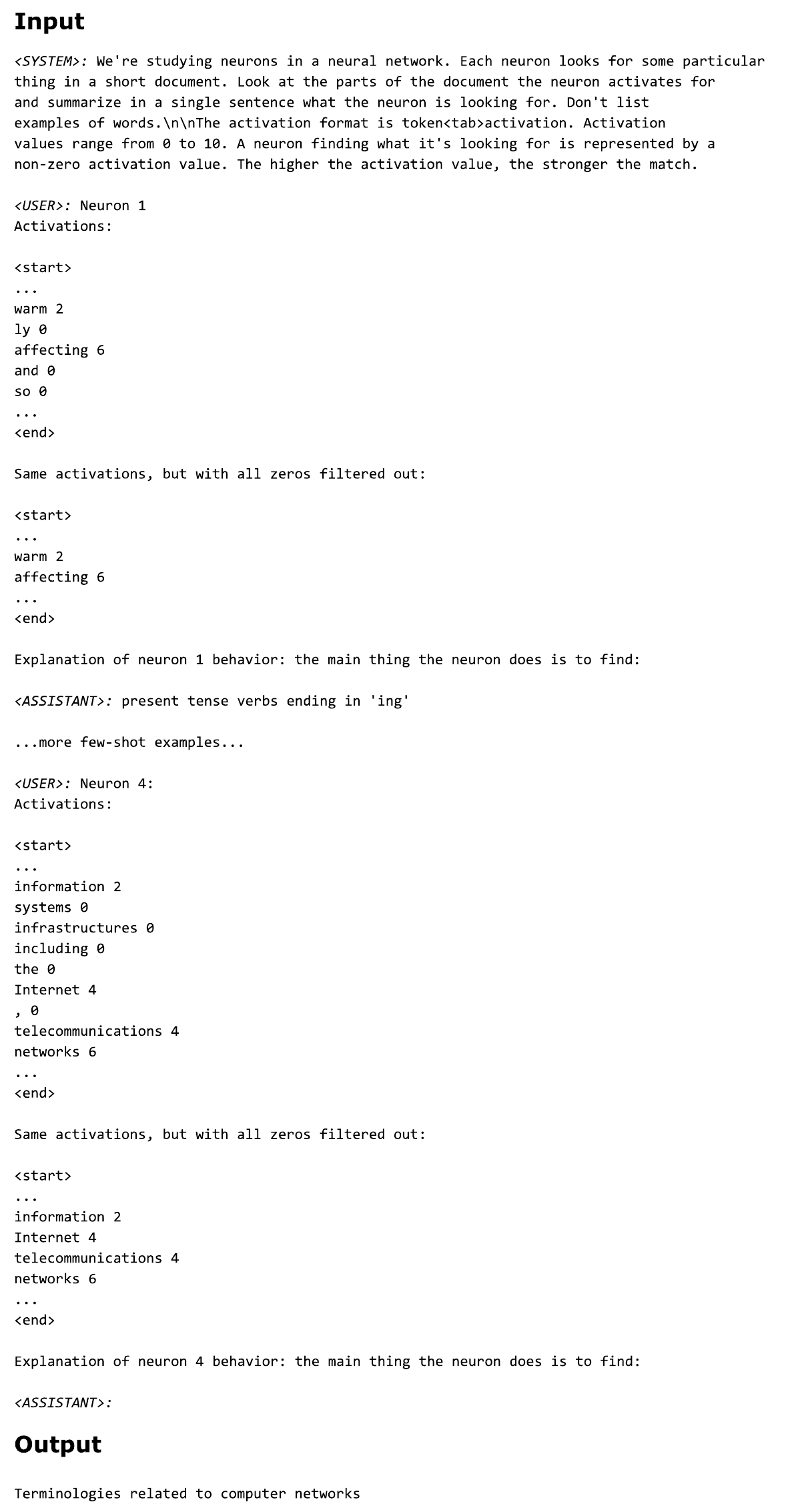}
  % \vspace{-20pt}
      \caption{Prompt and example input and output for generating a semantic expression for a given concept~\cite{bills2023language}. `\textbackslash t' is used as the separator between a token and an activation value.}
      \label{fig:automated_interpretability_prompt}
      \vspace{-10pt}
\end{figure*}

For LLM-based readability score~\cite{bills2023language}, we follow OpenAI's pipeline as illustrated in~\cite{bills2023language}. We show a detailed prompt for generating an explanation/semantic expression of a concept based on its activation in Fig.~\ref{fig:automated_interpretability_prompt}. We adopt this \href{https://github.com/openai/automated-interpretability/blob/main/neuron-explainer/neuron_explainer/explanations/simulator.py#L638}{adjusted algorithm} with \texttt{gpt-4-turbo-preview} as the simulator, due to new limitations in calculating logprobs on the input side. When extracting patterns that maximally activate a concept, we keep only the top 10 tokens with the largest activation or contribution to high-activation tokens.

\section{User study settings}
\label{app:user_instruction}

In our user study, we recruited 3 human labelers to evaluate the readability of 200 concepts. The human labelers possess a high school level of English proficiency, allowing for easy comprehension of the concepts. These labelers were selected from within our academic institution to ensure a consistent educational background, which is pertinent to the readability aspect of our study. To maintain the quality of labeling, we implemented a compensation structure that rewards the labelers based on the number of concepts they evaluate. This approach was designed to incentivize thorough and careful consideration of each concept.

During the study, labelers were required to complete their assessments within a five-minute window for each concept. This time constraint was established to simulate a realistic scenario in which users make quick judgments about concept readability. Each of the three labelers was presented with the same set of 200 concepts to ensure consistency in the evaluation process. 

Given input or output side tokens for a concept, each of our human labelers gives one readability score by simultaneously considering the three aspects, including semantic, grammatical or syntactic, and morphological information. More specifically, a concept is considered highly readable, if it is related to a specific topic such as computer systems (semantically interesting), is associated with a specific grammar or syntax (grammatically or syntactically interesting), or consists of tokens that share a similar structure or a form such as all being usable as suffixes for a certain token (morphologically interesting).

We demonstrate guidelines that were provided to the labelers. These guidelines were crafted to assist the labelers in their task and to standardize the evaluation criteria across all participants. The guidelines are as follows:

\textit{Welcome to the user study on evaluating the readability of concepts extracted from concept-based explanations. Your valuable insights will contribute to advancing our understanding of these explanations and improving their interpretability. Below are the instructions for scoring each concept:}

\textit{\textbf{Task Overview.}
You will be provided with a list of concepts, each comprising three parts:
\begin{itemize}
    \item Activation of this concept in 10 sentences, with each sentence containing 64 tokens.
    \item The 20 tokens that have the greatest impact on its activation value.
    \item The model's output of the 20 tokens with the highest logits after replacing hidden states with the direction of the concept.
\end{itemize}
For each concept, please provide two scores within the range of 
$[1, 2, 3, 4, 5]$, representing their perceived readability of the relevant information on the input and output sides.}

\textit{\textbf{Evaluation Criteria.} Please consider the following aspects when scoring each concept:
\begin{itemize}
    \item Semantic Information: Consider whether the concept is related to a specific topic, such as containing terms related to computer systems.
    \item Grammatical or syntactic information: Assess whether the concept is associated with specific grammar or syntax, such as being frequently activated with various copulas.
    \item Morphological Information: Consider whether the given tokens share a similar structure or form, such as all being usable as suffixes for a certain token.
\end{itemize}}

\textit{\textbf{Scoring Procedure.}
Please provide a score for the input side, reflecting the readability of tokens related to the concept in the input. Additionally, assign a score for the output side, indicating the readability of tokens related to the concept in the output. Your engagement in this scoring procedure will significantly contribute to the comprehensiveness of our study. Thank you for your participation!}

% \section{Hypothesis test on the experiments}
% \label{app:hypothesis_test}
% \begin{figure}[htbp]
%   \centering
%   % \vspace{-15pt}
%   \includegraphics[width=\linewidth]{figures/validity_relevance_pvalue.png}
%   % \vspace{-20pt}
%       \caption{Kendall $\tau$'s p-value of vadility results in Section~\ref{sec:main_results}.}
%       \label{fig:validity_relevance_pvalue}
%       \vspace{-10pt}
% \end{figure}

%% file: figures/taxonomy.tex
\begin{figure*}
    \centering
    \begin{forest} for tree={
        grow=east,
        growth parent anchor=east,
        parent anchor=east,
        child anchor=west,
        edge path={\noexpand\path[\forestoption{edge},->, >={latex}] 
             (!u.parent anchor) -- +(5pt,0pt) |- (.child anchor)
             \forestoption{edge label};}
    }
    [Concept Evaluation Metrics, root
        [Others, onode
            [Goodness:~\cite{hoffman2018metrics}, tnode]
            % [Simulatability:~\cite{hase2020evaluating}, tnode]
            [Importance:~\cite{ghorbani2019towards,chen2020concept,fel2023craft}, tnode]
            [Robustness:~\cite{kori2020abstracting,alvarez2018towards,sinha2023understanding}, tnode]
[Sensitivity/(In)stability/(In)consistency:~\cite{kim2018interpretability,alvarez2018towards,fel2023holistic,ghandeharioun2021dissect,kori2020abstracting,mikriukov2023evaluating,rosenfeld2021better}, tnode] 
            ]
        [Faithfulness, onode
            [Completeness:~\cite{yeh2020completeness,wang2023learning}, tnode]
            [Informativeness:~\cite{do2019theory}, tnode]
            [Reconstruction:~\cite{fel2023holistic}, tnode]
            [Fidelity:~\cite{fel2023holistic,sarkar2022framework,zhang2021invertible}, tnode] 
            [Faithfulness:~\cite{alvarez2018towards,hennigen2020intrinsic,sarkar2022framework,sun2023explain}, tnode] 
            ]
        [Readability, onode
            [Causality:~\cite{goyal2019explaining,wu2023causal}, tnode]
            % [Simulation and Score:, tnode]
            [Realism:~\cite{ghandeharioun2021dissect}, tnode]
        [Separability/Purity/Distincetness:~\cite{chen2020concept,ghandeharioun2021dissect,wang2023learning,do2019theory}, tnode] 
            [Alignment (with pre-defined concepts):~\cite{bau2017network,dalvi2022discovering,sarkar2022framework,na2019discovery,sajjad2022analyzing}, tnode]
            [Meaningfulness:~\cite{ghorbani2019towards}, tnode]
            [Sparsity/Complexity/Size:~\cite{rosenfeld2021better,lage2019evaluation}, tnode]
            ]
        ]
    \end{forest}
    \caption{Taxonomy of prior automatic metrics on concept-based explanation methods.}
    \label{fig:taxonomy}
\end{figure*}

%% file: tables/model_stat.tex
\begin{table}
    \centering
    \begin{tabular}{cccc}
    \hline
        Model & \#Layer & \#Param & \#Dimesion\\
    \hline
        GPT-2 (small) & 12 & 124M & 768\\
        % LLaMa-7b-chat & 32 & 5B & 4096\\
        Pythia-70M & 6 & 70M & 512\\
    \hline
    \end{tabular}
    \caption{Statistical model properties for subject models. \#Layer, \#Param, and \#Dimension represent the number of layers, parameters, and dimensions respectively.}
    \label{tab:model_property}
    \vspace{-10pt}
\end{table}
% source of the statistic 124M：https://openai.com/research/gpt-2-1-5b-release

%% file: acl.bbl
\begin{thebibliography}{80}
\expandafter\ifx\csname natexlab\endcsname\relax\def\natexlab#1{#1}\fi

\bibitem[{Adebayo et~al.(2018)Adebayo, Gilmer, Muelly, Goodfellow, Hardt, and Kim}]{adebayo2018sanity}
Julius Adebayo, Justin Gilmer, Michael Muelly, Ian Goodfellow, Moritz Hardt, and Been Kim. 2018.
\newblock Sanity checks for saliency maps.
\newblock \emph{Advances in neural information processing systems}, 31.

\bibitem[{Allen and Yen(2001)}]{allen2001introduction}
Mary~J Allen and Wendy~M Yen. 2001.
\newblock \emph{Introduction to measurement theory}.
\newblock Waveland Press.

\bibitem[{Alvarez~Melis and Jaakkola(2018)}]{alvarez2018towards}
David Alvarez~Melis and Tommi Jaakkola. 2018.
\newblock Towards robust interpretability with self-explaining neural networks.
\newblock \emph{Advances in neural information processing systems}, 31.

\bibitem[{Bau et~al.(2017)Bau, Zhou, Khosla, Oliva, and Torralba}]{bau2017network}
David Bau, Bolei Zhou, Aditya Khosla, Aude Oliva, and Antonio Torralba. 2017.
\newblock Network dissection: Quantifying interpretability of deep visual representations.
\newblock In \emph{Proceedings of the IEEE conference on computer vision and pattern recognition}, pages 6541--6549.

\bibitem[{Bhardwaj and Poria(2023)}]{bhardwaj2023red}
Rishabh Bhardwaj and Soujanya Poria. 2023.
\newblock Red-teaming large language models using chain of utterances for safety-alignment.
\newblock \emph{arXiv preprint arXiv:2308.09662}.

\bibitem[{Biderman et~al.(2023)Biderman, Schoelkopf, Anthony, Bradley, O’Brien, Hallahan, Khan, Purohit, Prashanth, Raff et~al.}]{biderman2023pythia}
Stella Biderman, Hailey Schoelkopf, Quentin~Gregory Anthony, Herbie Bradley, Kyle O’Brien, Eric Hallahan, Mohammad~Aflah Khan, Shivanshu Purohit, USVSN~Sai Prashanth, Edward Raff, et~al. 2023.
\newblock Pythia: A suite for analyzing large language models across training and scaling.
\newblock In \emph{International Conference on Machine Learning}, pages 2397--2430. PMLR.

\bibitem[{Bills et~al.(2023)Bills, Cammarata, Mossing, Tillman, Gao, Goh, Sutskever, Leike, Wu, and Saunders}]{bills2023language}
Steven Bills, Nick Cammarata, Dan Mossing, Henk Tillman, Leo Gao, Gabriel Goh, Ilya Sutskever, Jan Leike, Jeff Wu, and William Saunders. 2023.
\newblock Language models can explain neurons in language models.
\newblock \url{https://openaipublic.blob.core.windows.net/neuron-explainer/paper/index.html}.

\bibitem[{Bricken et~al.(2023)Bricken, Templeton, Batson, Chen, Jermyn, Conerly, Turner, Anil, Denison, Askell, Lasenby, Wu, Kravec, Schiefer, Maxwell, Joseph, Hatfield-Dodds, Tamkin, Nguyen, McLean, Burke, Hume, Carter, Henighan, and Olah}]{bricken2023monosemanticity}
Trenton Bricken, Adly Templeton, Joshua Batson, Brian Chen, Adam Jermyn, Tom Conerly, Nick Turner, Cem Anil, Carson Denison, Amanda Askell, Robert Lasenby, Yifan Wu, Shauna Kravec, Nicholas Schiefer, Tim Maxwell, Nicholas Joseph, Zac Hatfield-Dodds, Alex Tamkin, Karina Nguyen, Brayden McLean, Josiah~E Burke, Tristan Hume, Shan Carter, Tom Henighan, and Christopher Olah. 2023.
\newblock Towards monosemanticity: Decomposing language models with dictionary learning.
\newblock \emph{Transformer Circuits Thread}.
\newblock Https://transformer-circuits.pub/2023/monosemantic-features/index.html.

\bibitem[{Burger et~al.(2023)Burger, Chen, and Le}]{burger2023your}
Christopher Burger, Lingwei Chen, and Thai Le. 2023.
\newblock “are your explanations reliable?” investigating the stability of lime in explaining text classifiers by marrying xai and adversarial attack.
\newblock In \emph{Proceedings of the 2023 Conference on Empirical Methods in Natural Language Processing}, pages 12831--12844.

\bibitem[{Campbell and Fiske(1959)}]{campbell1959convergent}
Donald~T Campbell and Donald~W Fiske. 1959.
\newblock Convergent and discriminant validation by the multitrait-multimethod matrix.
\newblock \emph{Psychological bulletin}, 56(2):81.

\bibitem[{Chan et~al.(2022)Chan, Kong, and Liang}]{chan2022comparative}
Chun~Sik Chan, Huanqi Kong, and Guanqing Liang. 2022.
\newblock A comparative study of faithfulness metrics for model interpretability methods.
\newblock \emph{arXiv preprint arXiv:2204.05514}.

\bibitem[{Chen et~al.(2019{\natexlab{a}})Chen, Li, Tao, Barnett, Rudin, and Su}]{chen2019looks}
Chaofan Chen, Oscar Li, Daniel Tao, Alina Barnett, Cynthia Rudin, and Jonathan~K Su. 2019{\natexlab{a}}.
\newblock This looks like that: deep learning for interpretable image recognition.
\newblock \emph{Advances in neural information processing systems}, 32.

\bibitem[{Chen et~al.(2020)Chen, Bei, and Rudin}]{chen2020concept}
Zhi Chen, Yijie Bei, and Cynthia Rudin. 2020.
\newblock Concept whitening for interpretable image recognition.
\newblock \emph{Nature Machine Intelligence}, 2(12):772--782.

\bibitem[{Chen et~al.(2019{\natexlab{b}})Chen, Wang, Xie, Wu, Bu, Wang, and Chen}]{chen2019co}
Zhongxia Chen, Xiting Wang, Xing Xie, Tong Wu, Guoqing Bu, Yining Wang, and Enhong Chen. 2019{\natexlab{b}}.
\newblock Co-attentive multi-task learning for explainable recommendation.
\newblock In \emph{IJCAI}, volume 2019, pages 2137--2143.

\bibitem[{Clark et~al.(2021)Clark, August, Serrano, Haduong, Gururangan, and Smith}]{clark2021thats}
Elizabeth Clark, Tal August, Sofia Serrano, Nikita Haduong, Suchin Gururangan, and Noah~A. Smith. 2021.
\newblock \href {https://doi.org/10.18653/v1/2021.acl-long.565} {All that{'}s {`}human{'} is not gold: Evaluating human evaluation of generated text}.
\newblock In \emph{Proceedings of the 59th Annual Meeting of the Association for Computational Linguistics and the 11th International Joint Conference on Natural Language Processing (Volume 1: Long Papers)}, pages 7282--7296, Online. Association for Computational Linguistics.

\bibitem[{Colin et~al.(2022)Colin, Fel, Cad{\`e}ne, and Serre}]{colin2022cannot}
Julien Colin, Thomas Fel, R{\'e}mi Cad{\`e}ne, and Thomas Serre. 2022.
\newblock What i cannot predict, i do not understand: A human-centered evaluation framework for explainability methods.
\newblock \emph{Advances in Neural Information Processing Systems}, 35:2832--2845.

\bibitem[{Cronbach(1951)}]{cronbach1951coefficient}
Lee~J Cronbach. 1951.
\newblock Coefficient alpha and the internal structure of tests.
\newblock \emph{psychometrika}, 16(3):297--334.

\bibitem[{Cronbach and Meehl(1955)}]{cronbach1955construct}
Lee~J Cronbach and Paul~E Meehl. 1955.
\newblock Construct validity in psychological tests.
\newblock \emph{Psychological bulletin}, 52(4):281.

\bibitem[{Cunningham et~al.(2023)Cunningham, Ewart, Riggs, Huben, and Sharkey}]{cunningham2023sparse}
Hoagy Cunningham, Aidan Ewart, Logan Riggs, Robert Huben, and Lee Sharkey. 2023.
\newblock Sparse autoencoders find highly interpretable features in language models.
\newblock \emph{arXiv preprint arXiv:2309.08600}.

\bibitem[{Dalvi et~al.(2022)Dalvi, Khan, Alam, Durrani, Xu, and Sajjad}]{dalvi2022discovering}
Fahim Dalvi, Abdul~Rafae Khan, Firoj Alam, Nadir Durrani, Jia Xu, and Hassan Sajjad. 2022.
\newblock Discovering latent concepts learned in bert.
\newblock \emph{arXiv preprint arXiv:2205.07237}.

\bibitem[{Do and Tran(2019)}]{do2019theory}
Kien Do and Truyen Tran. 2019.
\newblock Theory and evaluation metrics for learning disentangled representations.
\newblock \emph{arXiv preprint arXiv:1908.09961}.

\bibitem[{Elhage et~al.(2022)Elhage, Hume, Olsson, Schiefer, Henighan, Kravec, Hatfield-Dodds, Lasenby, Drain, Chen et~al.}]{elhage2022toy}
Nelson Elhage, Tristan Hume, Catherine Olsson, Nicholas Schiefer, Tom Henighan, Shauna Kravec, Zac Hatfield-Dodds, Robert Lasenby, Dawn Drain, Carol Chen, et~al. 2022.
\newblock Toy models of superposition.
\newblock \emph{arXiv preprint arXiv:2209.10652}.

\bibitem[{Fel et~al.(2023{\natexlab{a}})Fel, Boutin, Moayeri, Cad{\`e}ne, Bethune, Chalvidal, Serre et~al.}]{fel2023holistic}
Thomas Fel, Victor Boutin, Mazda Moayeri, R{\'e}mi Cad{\`e}ne, Louis Bethune, Mathieu Chalvidal, Thomas Serre, et~al. 2023{\natexlab{a}}.
\newblock A holistic approach to unifying automatic concept extraction and concept importance estimation.
\newblock \emph{arXiv preprint arXiv:2306.07304}.

\bibitem[{Fel et~al.(2023{\natexlab{b}})Fel, Picard, Bethune, Boissin, Vigouroux, Colin, Cad{\`e}ne, and Serre}]{fel2023craft}
Thomas Fel, Agustin Picard, Louis Bethune, Thibaut Boissin, David Vigouroux, Julien Colin, R{\'e}mi Cad{\`e}ne, and Thomas Serre. 2023{\natexlab{b}}.
\newblock Craft: Concept recursive activation factorization for explainability.
\newblock In \emph{Proceedings of the IEEE/CVF Conference on Computer Vision and Pattern Recognition}, pages 2711--2721.

\bibitem[{Galton(1877)}]{galton1877typical}
Francis Galton. 1877.
\newblock \href {https://doi.org/10.1038/015492a0} {Typical laws of heredity 1}.
\newblock \emph{Nature}, 15(388):492--495.

\bibitem[{Gao et~al.(2019)Gao, Wang, Wang, and Xie}]{gao2019explainable}
Jingyue Gao, Xiting Wang, Yasha Wang, and Xing Xie. 2019.
\newblock Explainable recommendation through attentive multi-view learning.
\newblock In \emph{Proceedings of the AAAI Conference on Artificial Intelligence}, volume~33, pages 3622--3629.

\bibitem[{Gao et~al.(2020)Gao, Biderman, Black, Golding, Hoppe, Foster, Phang, He, Thite, Nabeshima et~al.}]{gao2020pile}
Leo Gao, Stella Biderman, Sid Black, Laurence Golding, Travis Hoppe, Charles Foster, Jason Phang, Horace He, Anish Thite, Noa Nabeshima, et~al. 2020.
\newblock The pile: An 800gb dataset of diverse text for language modeling.
\newblock \emph{arXiv preprint arXiv:2101.00027}.

\bibitem[{Ghandeharioun et~al.(2021)Ghandeharioun, Kim, Li, Jou, Eoff, and Picard}]{ghandeharioun2021dissect}
Asma Ghandeharioun, Been Kim, Chun-Liang Li, Brendan Jou, Brian Eoff, and Rosalind~W Picard. 2021.
\newblock Dissect: Disentangled simultaneous explanations via concept traversals.
\newblock \emph{arXiv preprint arXiv:2105.15164}.

\bibitem[{Ghorbani et~al.(2019)Ghorbani, Wexler, Zou, and Kim}]{ghorbani2019towards}
Amirata Ghorbani, James Wexler, James~Y Zou, and Been Kim. 2019.
\newblock Towards automatic concept-based explanations.
\newblock \emph{Advances in neural information processing systems}, 32.

\bibitem[{Goyal et~al.(2019)Goyal, Feder, Shalit, and Kim}]{goyal2019explaining}
Yash Goyal, Amir Feder, Uri Shalit, and Been Kim. 2019.
\newblock Explaining classifiers with causal concept effect (cace).
\newblock \emph{arXiv preprint arXiv:1907.07165}.

\bibitem[{Guan et~al.(2019)Guan, Wang, Zhang, Chen, He, and Xie}]{guan2019towards}
Chaoyu Guan, Xiting Wang, Quanshi Zhang, Runjin Chen, Di~He, and Xing Xie. 2019.
\newblock Towards a deep and unified understanding of deep neural models in nlp.
\newblock In \emph{International conference on machine learning}, pages 2454--2463. PMLR.

\bibitem[{Hennigen et~al.(2020)Hennigen, Williams, and Cotterell}]{hennigen2020intrinsic}
Lucas~Torroba Hennigen, Adina Williams, and Ryan Cotterell. 2020.
\newblock Intrinsic probing through dimension selection.
\newblock \emph{arXiv preprint arXiv:2010.02812}.

\bibitem[{Hoffman et~al.(2018)Hoffman, Mueller, Klein, and Litman}]{hoffman2018metrics}
Robert~R Hoffman, Shane~T Mueller, Gary Klein, and Jordan Litman. 2018.
\newblock Metrics for explainable ai: Challenges and prospects.
\newblock \emph{arXiv preprint arXiv:1812.04608}.

\bibitem[{Howcroft et~al.(2020)Howcroft, Belz, Clinciu, Gkatzia, Hasan, Mahamood, Mille, van Miltenburg, Santhanam, and Rieser}]{howcroft2020twenty}
David~M. Howcroft, Anya Belz, Miruna-Adriana Clinciu, Dimitra Gkatzia, Sadid~A. Hasan, Saad Mahamood, Simon Mille, Emiel van Miltenburg, Sashank Santhanam, and Verena Rieser. 2020.
\newblock \href {https://doi.org/10.18653/v1/2020.inlg-1.23} {Twenty years of confusion in human evaluation: {NLG} needs evaluation sheets and standardised definitions}.
\newblock In \emph{Proceedings of the 13th International Conference on Natural Language Generation}, pages 169--182, Dublin, Ireland. Association for Computational Linguistics.

\bibitem[{Jacovi and Goldberg(2020)}]{jacovi2020towards}
Alon Jacovi and Yoav Goldberg. 2020.
\newblock Towards faithfully interpretable nlp systems: How should we define and evaluate faithfulness?
\newblock \emph{arXiv preprint arXiv:2004.03685}.

\bibitem[{Jin et~al.(2022)Jin, Wang, Yang, Sun, Wang, Liao, and Xie}]{jin2022towards}
Yiqiao Jin, Xiting Wang, Ruichao Yang, Yizhou Sun, Wei Wang, Hao Liao, and Xing Xie. 2022.
\newblock Towards fine-grained reasoning for fake news detection.
\newblock In \emph{Proceedings of the AAAI Conference on Artificial Intelligence}, volume~36, pages 5746--5754.

\bibitem[{Kendall(1938)}]{kendall1938new}
Maurice~G Kendall. 1938.
\newblock A new measure of rank correlation.
\newblock \emph{Biometrika}, 30(1/2):81--93.

\bibitem[{Kim et~al.(2018)Kim, Wattenberg, Gilmer, Cai, Wexler, Viegas et~al.}]{kim2018interpretability}
Been Kim, Martin Wattenberg, Justin Gilmer, Carrie Cai, James Wexler, Fernanda Viegas, et~al. 2018.
\newblock Interpretability beyond feature attribution: Quantitative testing with concept activation vectors (tcav).
\newblock In \emph{International conference on machine learning}, pages 2668--2677. PMLR.

\bibitem[{Koh et~al.(2020)Koh, Nguyen, Tang, Mussmann, Pierson, Kim, and Liang}]{koh2020concept}
Pang~Wei Koh, Thao Nguyen, Yew~Siang Tang, Stephen Mussmann, Emma Pierson, Been Kim, and Percy Liang. 2020.
\newblock Concept bottleneck models.
\newblock In \emph{International conference on machine learning}, pages 5338--5348. PMLR.

\bibitem[{Kori et~al.(2020)Kori, Natekar, Krishnamurthi, and Srinivasan}]{kori2020abstracting}
Avinash Kori, Parth Natekar, Ganapathy Krishnamurthi, and Balaji Srinivasan. 2020.
\newblock Abstracting deep neural networks into concept graphs for concept level interpretability.
\newblock \emph{arXiv preprint arXiv:2008.06457}.

\bibitem[{Lage et~al.(2019)Lage, Chen, He, Narayanan, Kim, Gershman, and Doshi-Velez}]{lage2019evaluation}
Isaac Lage, Emily Chen, Jeffrey He, Menaka Narayanan, Been Kim, Sam Gershman, and Finale Doshi-Velez. 2019.
\newblock An evaluation of the human-interpretability of explanation.
\newblock \emph{arXiv preprint arXiv:1902.00006}.

\bibitem[{Lee et~al.(2023)Lee, Lanza, and Wermter}]{lee2023neural}
Jae~Hee Lee, Sergio Lanza, and Stefan Wermter. 2023.
\newblock From neural activations to concepts: A survey on explaining concepts in neural networks.
\newblock \emph{arXiv preprint arXiv:2310.11884}.

\bibitem[{Lee et~al.(2022)Lee, Wang, Han, Yi, Xie, and Cha}]{lee2022self}
Seungeon Lee, Xiting Wang, Sungwon Han, Xiaoyuan Yi, Xing Xie, and Meeyoung Cha. 2022.
\newblock Self-explaining deep models with logic rule reasoning.
\newblock \emph{Advances in Neural Information Processing Systems}, 35:3203--3216.

\bibitem[{Li et~al.(2023)Li, Wang, Zheng, and Zhang}]{li2023loogle}
Jiaqi Li, Mengmeng Wang, Zilong Zheng, and Muhan Zhang. 2023.
\newblock Loogle: Can long-context language models understand long contexts?
\newblock \emph{arXiv preprint arXiv:2311.04939}.

\bibitem[{Li et~al.(2022)Li, Wang, Yang, Wu, Zhang, Liu, Sun, Zhang, and Liu}]{li2022unified}
Zhen Li, Xiting Wang, Weikai Yang, Jing Wu, Zhengyan Zhang, Zhiyuan Liu, Maosong Sun, Hui Zhang, and Shixia Liu. 2022.
\newblock A unified understanding of deep nlp models for text classification.
\newblock \emph{IEEE Transactions on Visualization and Computer Graphics}, 28(12):4980--4994.

\bibitem[{Lundberg and Lee(2017)}]{lundberg2017unified}
Scott~M Lundberg and Su-In Lee. 2017.
\newblock A unified approach to interpreting model predictions.
\newblock \emph{Advances in neural information processing systems}, 30.

\bibitem[{McCarthy and Prince(1995)}]{mccarthy1995faithfulness}
John~J McCarthy and Alan Prince. 1995.
\newblock Faithfulness and reduplicative identity.
\newblock \emph{Linguistics Department Faculty Publication Series}, page~10.

\bibitem[{Mikriukov et~al.(2023)Mikriukov, Schwalbe, Hellert, and Bade}]{mikriukov2023evaluating}
Georgii Mikriukov, Gesina Schwalbe, Christian Hellert, and Korinna Bade. 2023.
\newblock Evaluating the stability of semantic concept representations in cnns for robust explainability.
\newblock \emph{arXiv preprint arXiv:2304.14864}.

\bibitem[{Mimno et~al.(2011)Mimno, Wallach, Talley, Leenders, and McCallum}]{mimno2011optimizing}
David Mimno, Hanna Wallach, Edmund Talley, Miriam Leenders, and Andrew McCallum. 2011.
\newblock Optimizing semantic coherence in topic models.
\newblock In \emph{Proceedings of the 2011 conference on empirical methods in natural language processing}, pages 262--272.

\bibitem[{Na et~al.(2019)Na, Choe, Lee, and Kim}]{na2019discovery}
Seil Na, Yo~Joong Choe, Dong-Hyun Lee, and Gunhee Kim. 2019.
\newblock Discovery of natural language concepts in individual units of cnns.
\newblock \emph{arXiv preprint arXiv:1902.07249}.

\bibitem[{Newman et~al.(2009)Newman, Karimi, and Cavedon}]{newman2009external}
David Newman, Sarvnaz Karimi, and Lawrence Cavedon. 2009.
\newblock External evaluation of topic models.
\newblock In \emph{Proceedings of the 14th Australasian Document Computing Symposium}, pages 1--8. University of Sydney.

\bibitem[{Newman et~al.(2010)Newman, Lau, Grieser, and Baldwin}]{newman2010automatic}
David Newman, Jey~Han Lau, Karl Grieser, and Timothy Baldwin. 2010.
\newblock Automatic evaluation of topic coherence.
\newblock In \emph{Human language technologies: The 2010 annual conference of the North American chapter of the association for computational linguistics}, pages 100--108.

\bibitem[{Nunnally and Bernstein(1994)}]{nunnally1994psychometric}
Jum~C Nunnally and Ira~H Bernstein. 1994.
\newblock Psychometric theory new york.
\newblock \emph{NY: McGraw-Hill}.

\bibitem[{Radford et~al.(2021)Radford, Kim, Hallacy, Ramesh, Goh, Agarwal, Sastry, Askell, Mishkin, Clark et~al.}]{radford2021learning}
Alec Radford, Jong~Wook Kim, Chris Hallacy, Aditya Ramesh, Gabriel Goh, Sandhini Agarwal, Girish Sastry, Amanda Askell, Pamela Mishkin, Jack Clark, et~al. 2021.
\newblock Learning transferable visual models from natural language supervision.
\newblock In \emph{International conference on machine learning}, pages 8748--8763. PMLR.

\bibitem[{Radford et~al.(2019)Radford, Wu, Child, Luan, Amodei, Sutskever et~al.}]{radford2019language}
Alec Radford, Jeffrey Wu, Rewon Child, David Luan, Dario Amodei, Ilya Sutskever, et~al. 2019.
\newblock Language models are unsupervised multitask learners.
\newblock \emph{OpenAI blog}, 1(8):9.

\bibitem[{Ribeiro et~al.(2016)Ribeiro, Singh, and Guestrin}]{ribeiro2016should}
Marco~Tulio Ribeiro, Sameer Singh, and Carlos Guestrin. 2016.
\newblock " why should i trust you?" explaining the predictions of any classifier.
\newblock In \emph{Proceedings of the 22nd ACM SIGKDD international conference on knowledge discovery and data mining}, pages 1135--1144.

\bibitem[{Rosenfeld(2021)}]{rosenfeld2021better}
Avi Rosenfeld. 2021.
\newblock Better metrics for evaluating explainable artificial intelligence.
\newblock In \emph{Proceedings of the 20th international conference on autonomous agents and multiagent systems}, pages 45--50.

\bibitem[{Sajjad et~al.(2022)Sajjad, Durrani, Dalvi, Alam, Khan, and Xu}]{sajjad2022analyzing}
Hassan Sajjad, Nadir Durrani, Fahim Dalvi, Firoj Alam, Abdul~Rafae Khan, and Jia Xu. 2022.
\newblock Analyzing encoded concepts in transformer language models.
\newblock \emph{arXiv preprint arXiv:2206.13289}.

\bibitem[{Sarkar et~al.(2022)Sarkar, Vijaykeerthy, Sarkar, and Balasubramanian}]{sarkar2022framework}
Anirban Sarkar, Deepak Vijaykeerthy, Anindya Sarkar, and Vineeth~N Balasubramanian. 2022.
\newblock A framework for learning ante-hoc explainable models via concepts.
\newblock In \emph{Proceedings of the IEEE/CVF Conference on Computer Vision and Pattern Recognition}, pages 10286--10295.

\bibitem[{Schwab and Karlen(2019)}]{schwab2019cxplain}
Patrick Schwab and Walter Karlen. 2019.
\newblock Cxplain: Causal explanations for model interpretation under uncertainty.
\newblock \emph{Advances in neural information processing systems}, 32.

\bibitem[{Simonyan and Zisserman(2014)}]{simonyan2014very}
Karen Simonyan and Andrew Zisserman. 2014.
\newblock Very deep convolutional networks for large-scale image recognition.
\newblock \emph{arXiv preprint arXiv:1409.1556}.

\bibitem[{Singh et~al.(2023)Singh, Hsu, Antonello, Jain, Huth, Yu, and Gao}]{singh2023explaining}
Chandan Singh, Aliyah~R Hsu, Richard Antonello, Shailee Jain, Alexander~G Huth, Bin Yu, and Jianfeng Gao. 2023.
\newblock Explaining black box text modules in natural language with language models.
\newblock \emph{arXiv preprint arXiv:2305.09863}.

\bibitem[{Sinha et~al.(2023)Sinha, Huai, Sun, and Zhang}]{sinha2023understanding}
Sanchit Sinha, Mengdi Huai, Jianhui Sun, and Aidong Zhang. 2023.
\newblock Understanding and enhancing robustness of concept-based models.
\newblock In \emph{Proceedings of the AAAI Conference on Artificial Intelligence}, volume~37, pages 15127--15135.

\bibitem[{Spearman(1961)}]{spearman1961proof}
C.~Spearman. 1961.
\newblock \href {https://doi.org/10.2307/1412159} {The proof and measurement of association between two things}.
\newblock \emph{The American Journal of Psychology}, 15(1):72--101.

\bibitem[{Sun et~al.(2023)Sun, Ma, Yuan, and Wang}]{sun2023explain}
Ao~Sun, Pingchuan Ma, Yuanyuan Yuan, and Shuai Wang. 2023.
\newblock Explain any concept: Segment anything meets concept-based explanation.
\newblock \emph{arXiv preprint arXiv:2305.10289}.

\bibitem[{Sundararajan et~al.(2017)Sundararajan, Taly, and Yan}]{sundararajan2017axiomatic}
Mukund Sundararajan, Ankur Taly, and Qiqi Yan. 2017.
\newblock Axiomatic attribution for deep networks.
\newblock In \emph{International conference on machine learning}, pages 3319--3328. PMLR.

\bibitem[{Wang et~al.(2023{\natexlab{a}})Wang, Li, Nakashima, and Nagahara}]{wang2023learning}
Bowen Wang, Liangzhi Li, Yuta Nakashima, and Hajime Nagahara. 2023{\natexlab{a}}.
\newblock Learning bottleneck concepts in image classification.
\newblock In \emph{Proceedings of the IEEE/CVF Conference on Computer Vision and Pattern Recognition}, pages 10962--10971.

\bibitem[{Wang et~al.(2023{\natexlab{b}})Wang, Zhang, Shen, and Kong}]{wang2023densecl}
Xinlong Wang, Rufeng Zhang, Chunhua Shen, and Tao Kong. 2023{\natexlab{b}}.
\newblock Densecl: A simple framework for self-supervised dense visual pre-training.
\newblock \emph{Visual Informatics}, 7(1):30--40.

\bibitem[{Wang et~al.(2023{\natexlab{c}})Wang, Jiang, Hernandez-Orallo, Sun, Stillwell, Luo, and Xie}]{wang2023evaluating}
Xiting Wang, Liming Jiang, Jose Hernandez-Orallo, Luning Sun, David Stillwell, Fang Luo, and Xing Xie. 2023{\natexlab{c}}.
\newblock Evaluating general-purpose ai with psychometrics.
\newblock \emph{arXiv preprint arXiv:2310.16379}.

\bibitem[{Wang et~al.(2022)Wang, Liu, Wang, Wu, Fu, and Xie}]{wang2022multi}
Xiting Wang, Kunpeng Liu, Dongjie Wang, Le~Wu, Yanjie Fu, and Xing Xie. 2022.
\newblock Multi-level recommendation reasoning over knowledge graphs with reinforcement learning.
\newblock In \emph{Proceedings of the ACM Web Conference 2022}, pages 2098--2108.

\bibitem[{Wu et~al.(2023{\natexlab{a}})Wu, Wang, Lian, Xie, and Chen}]{wu2023causality}
Chenwang Wu, Xiting Wang, Defu Lian, Xing Xie, and Enhong Chen. 2023{\natexlab{a}}.
\newblock A causality inspired framework for model interpretation.
\newblock In \emph{Proceedings of the 29th ACM SIGKDD Conference on Knowledge Discovery and Data Mining}, pages 2731--2741.

\bibitem[{Wu et~al.(2023{\natexlab{b}})Wu, D’Oosterlinck, Geiger, Zur, and Potts}]{wu2023causal}
Zhengxuan Wu, Karel D’Oosterlinck, Atticus Geiger, Amir Zur, and Christopher Potts. 2023{\natexlab{b}}.
\newblock Causal proxy models for concept-based model explanations.
\newblock In \emph{International conference on machine learning}, pages 37313--37334. PMLR.

\bibitem[{Xiao et~al.(2023)Xiao, Zhang, Lai, and Liao}]{xiao2023evaluating}
Ziang Xiao, Susu Zhang, Vivian Lai, and Q~Vera Liao. 2023.
\newblock Evaluating evaluation metrics: A framework for analyzing nlg evaluation metrics using measurement theory.
\newblock In \emph{Proceedings of the 2023 Conference on Empirical Methods in Natural Language Processing}, pages 10967--10982.

\bibitem[{Xu et~al.(2024)Xu, Huang, Wang, Wu, Yao, and Xie}]{xu2024uncovering}
Zhihao Xu, Ruixuan Huang, Xiting Wang, Fangzhao Wu, Jing Yao, and Xing Xie. 2024.
\newblock Uncovering safety risks in open-source llms through concept activation vector.
\newblock \emph{Advances in Neural Information Processing Systems}.

\bibitem[{Yang et~al.(2022)Yang, Wang, Jin, Li, Lian, and Xie}]{yang2022reinforcement}
Ruichao Yang, Xiting Wang, Yiqiao Jin, Chaozhuo Li, Jianxun Lian, and Xing Xie. 2022.
\newblock Reinforcement subgraph reasoning for fake news detection.
\newblock In \emph{Proceedings of the 28th ACM SIGKDD Conference on Knowledge Discovery and Data Mining}, pages 2253--2262.

\bibitem[{Yang et~al.(2024)Yang, Liu, Wang, and Liu}]{yang2024foundation}
Weikai Yang, Mengchen Liu, Zheng Wang, and Shixia Liu. 2024.
\newblock Foundation models meet visualizations: Challenges and opportunities.
\newblock \emph{Computational Visual Media}, pages 1--26.

\bibitem[{Yeh et~al.(2020)Yeh, Kim, Arik, Li, Pfister, and Ravikumar}]{yeh2020completeness}
Chih-Kuan Yeh, Been Kim, Sercan Arik, Chun-Liang Li, Tomas Pfister, and Pradeep Ravikumar. 2020.
\newblock On completeness-aware concept-based explanations in deep neural networks.
\newblock \emph{Advances in neural information processing systems}, 33:20554--20565.

\bibitem[{Zhang et~al.(2024)Zhang, Wang, Ao, and He}]{zhang2024distillation}
Hanyu Zhang, Xiting Wang, Xiang Ao, and Qing He. 2024.
\newblock Distillation with explanations from large language models.
\newblock In \emph{Proceedings of the 2024 Joint International Conference on Computational Linguistics, Language Resources and Evaluation (LREC-COLING 2024)}, pages 5018--5028.

\bibitem[{Zhang et~al.(2021)Zhang, Madumal, Miller, Ehinger, and Rubinstein}]{zhang2021invertible}
Ruihan Zhang, Prashan Madumal, Tim Miller, Krista~A Ehinger, and Benjamin~IP Rubinstein. 2021.
\newblock Invertible concept-based explanations for cnn models with non-negative concept activation vectors.
\newblock In \emph{Proceedings of the AAAI Conference on Artificial Intelligence}, pages 11682--11690.

\bibitem[{Zou et~al.(2023)Zou, Phan, Chen, Campbell, Guo, Ren, Pan, Yin, Mazeika, Dombrowski et~al.}]{zou2023representation}
Andy Zou, Long Phan, Sarah Chen, James Campbell, Phillip Guo, Richard Ren, Alexander Pan, Xuwang Yin, Mantas Mazeika, Ann-Kathrin Dombrowski, et~al. 2023.
\newblock Representation engineering: A top-down approach to ai transparency.
\newblock \emph{arXiv preprint arXiv:2310.01405}.

\end{thebibliography}
